\documentclass[10pt,twocolumn,letterpaper]{article}

\usepackage[pagenumbers]{cvpr} 

\usepackage[dvipsnames]{xcolor}

\definecolor{cvprblue}{rgb}{0.21,0.49,0.74}
\usepackage[pagebackref,breaklinks,colorlinks,citecolor=cvprblue]{hyperref}
\usepackage{multirow}
\usepackage{makecell}
\usepackage{tabularray}
\usepackage{colortbl}
\usepackage{graphicx}
\usepackage{bbding}
\usepackage{pifont}

\usepackage[accsupp]{axessibility}

\def\paperID{11674} 
\def\confName{CVPR}
\def\confYear{2024}

\title{LiDAR4D: Dynamic Neural Fields for Novel Space-time View LiDAR Synthesis}

\author{Zehan Zheng,~ Fan Lu,~ Weiyi Xue,~ Guang Chen$^{\dagger}$,~ Changjun Jiang\\
Tongji University\\
{\tt\small \{zhengzehan, lufan, xwy, guangchen, cjjiang\}@tongji.edu.cn}
}

\begin{document}
\maketitle

\renewcommand{\thefootnote}{}
\footnote{$^\dagger$ Corresponding author.}

\vspace{-0.5cm}

\begin{abstract}
Although neural radiance fields (NeRFs) have achieved triumphs in image novel view synthesis (NVS), LiDAR NVS remains largely unexplored. Previous LiDAR NVS methods employ a simple shift from image NVS methods while ignoring the dynamic nature and the large-scale reconstruction problem of LiDAR point clouds. In light of this, we propose \textbf{LiDAR4D}, a differentiable LiDAR-only framework for novel space-time LiDAR view synthesis. In consideration of the sparsity and large-scale characteristics, we design a 4D hybrid representation combined with multi-planar and grid features to achieve effective reconstruction in a coarse-to-fine manner. Furthermore, we introduce geometric constraints derived from point clouds to improve temporal consistency. For the realistic synthesis of LiDAR point clouds, we incorporate the global optimization of ray-drop probability to preserve cross-region patterns. Extensive experiments on KITTI-360 and NuScenes datasets demonstrate the superiority of our method in accomplishing geometry-aware and time-consistent dynamic reconstruction. Codes are available at \url{https://github.com/ispc-lab/LiDAR4D}.

\end{abstract}    
\section{Introduction}
\label{sec:intro}

Dynamic scene reconstruction is of crucial importance across various fields such as AR/VR, robotics and autonomous driving. Existing advanced methods in computer vision enable high-fidelity 3D scene reconstruction and novel view synthesis (NVS), which can further serve a wide range of downstream tasks and applications. For instance, we could reconstruct driving scenarios directly from collected sensor logs, allowing for scene replay and novel data generation~\cite{yang2023unisim}. It shows great potential for boosting data diversity, forming data closed-loop, and improving the generalizability of the autonomous driving system.

\begin{figure}[t]
\centering
  \includegraphics[width=0.47\textwidth]{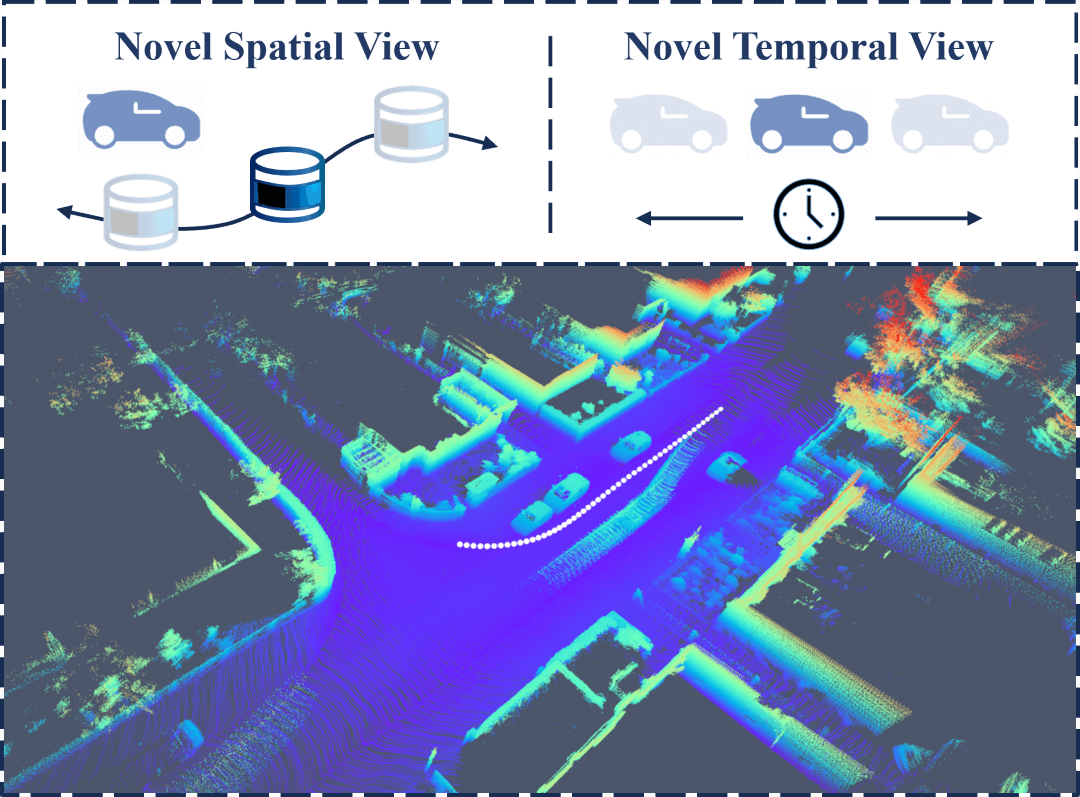}
  \caption{\textbf{Dynamic scenes of LiDAR point clouds in autonomous driving.} Large-scale vehicle motion poses a significant challenge for dynamic reconstruction and novel space-time view synthesis. White dots indicate the ego-car trajectory. }
  \label{fig:dynamic scenes}
\vspace{-0.2cm}
\end{figure}  

However, the majority of current research focuses on novel view synthesis for cameras, while other sensors such as LiDAR remain largely unexplored. Similar to camera images, LiDAR point clouds are also partial observations of the scene that vary across different locations and views. The reconstruction faces considerable challenges due to the sparsity, discontinuity and occlusion of LiDAR point clouds. Furthermore, as illustrated in \Cref{fig:dynamic scenes}, dynamic scenarios combine novel spatial view and temporal view synthesis simultaneously. Meanwhile, the large motion of dynamic objects makes it difficult to align and reconstruct.

Traditional LiDAR-based 3D scene reconstruction techniques aggregate multiple sparse point cloud frames directly in the world coordinate system~\cite{li2023pcgen} and further convert them into explicit surface representations such as triangular meshes~\cite{manivasagam2020lidarsim}. Subsequently, the intersection of the LiDAR beams with the mesh surface can be calculated by performing ray-casting to render novel-view LiDAR point clouds.

Nevertheless, high-quality surface reconstruction of complex large-scale scenes is challenging to accomplish, which may lead to significant geometric errors. Furthermore, the aforementioned explicit reconstruction method is limited to static scenes and struggles to accurately model the intensity or ray-drop characteristics of actual LiDAR points.

Neural radiance fields~\cite{mildenhall2021nerf} implicitly reconstruct the scene and generate novel-view data through volume rendering in a continuous representation space, which also offers an alternative solution for LiDAR reconstruction. Consequently, the most recent researches~\cite{zhang2023nerflidar,huang2023nfl,tao2023lidarnerf} are shifting their attention towards the novel view synthesis of LiDAR. NeRF-LiDAR~\cite{zhang2023nerflidar} integrates image and point cloud modalities for LiDAR synthesis, whereas LiDAR-only methods like LiDAR-NeRF~\cite{tao2023lidarnerf} and NFL~\cite{huang2023nfl} explore the possibility of LiDAR reconstruction and generation without RGB images. Most prior methods directly apply the image NVS pipeline to LiDAR point clouds. However, LiDAR point clouds are inherently different from 2D images, which poses challenges for current LiDAR NVS methods to achieve high-quality reconstruction: (1) previous methods are limited to static scenes, ignoring the dynamic nature of autonomous driving scenarios; (2) the vast scale and high sparsity of LiDAR point clouds pose higher demands on the representations; and (3) intensity and ray-drop characteristics modeling are required for synthesis realism.

To overcome the aforementioned limitations, we propose LiDAR4D, shedding light on three pivotal insights to elevate the current LiDAR NVS pipeline. To tackle the dynamic objects, we introduce geometric constraints derived from point clouds and aggregate multi-frame dynamic features for temporal consistency. Regarding compact large-scale scene reconstruction, we design a coarse-to-fine hybrid representation combined with multi-planar and grid features to reconstruct the smooth geometry and high-frequency intensity. Additionally, we employ global optimization to preserve patterns across regions for ray-drop probability refinement. Therefore, LiDAR4D is capable of achieving geometry-aware and time-consistent reconstruction under large-scale dynamic scenarios.

We evaluate our method on diverse dynamic scenarios of KITTI-360~\cite{liao2022kitti} and NuScenes~\cite{caesar2020nuscenes} autonomous driving datasets. With comprehensive experiments, LiDAR4D significantly outperforms previous state-of-the-art NeRF-based implicit approaches and explicit reconstruction methods. In comparison to LiDAR-NeRF~\cite{tao2023lidarnerf}, we achieve 24.3\% and 24.2\% reduction in CD error on KITTI-360 dataset and NuScenes dataset, respectively. Similar leadership exists for other metrics of range depth and intensity.

In summary, our main contributions are three-fold:
\begin{itemize} 
\item We propose LiDAR4D, a differentiable LiDAR-only framework for novel space-time LiDAR view synthesis, which reconstructs dynamic driving scenarios and generates realistic LiDAR point clouds end-to-end.

\vspace{0.05cm}

\item We introduce 4D hybrid neural representations and motion priors derived from point clouds for geometry-aware and time-consistent large-scale scene reconstruction.

\vspace{0.05cm}

\item Comprehensive experiments demonstrate the state-of-the-art performance of LiDAR4D in challenging dynamic scene reconstruction and novel view synthesis. 
\end{itemize}
\section{Related Work}
\label{sec: related work}

\begin{figure*}[ht]
\centering
  \includegraphics[width=0.95\textwidth]{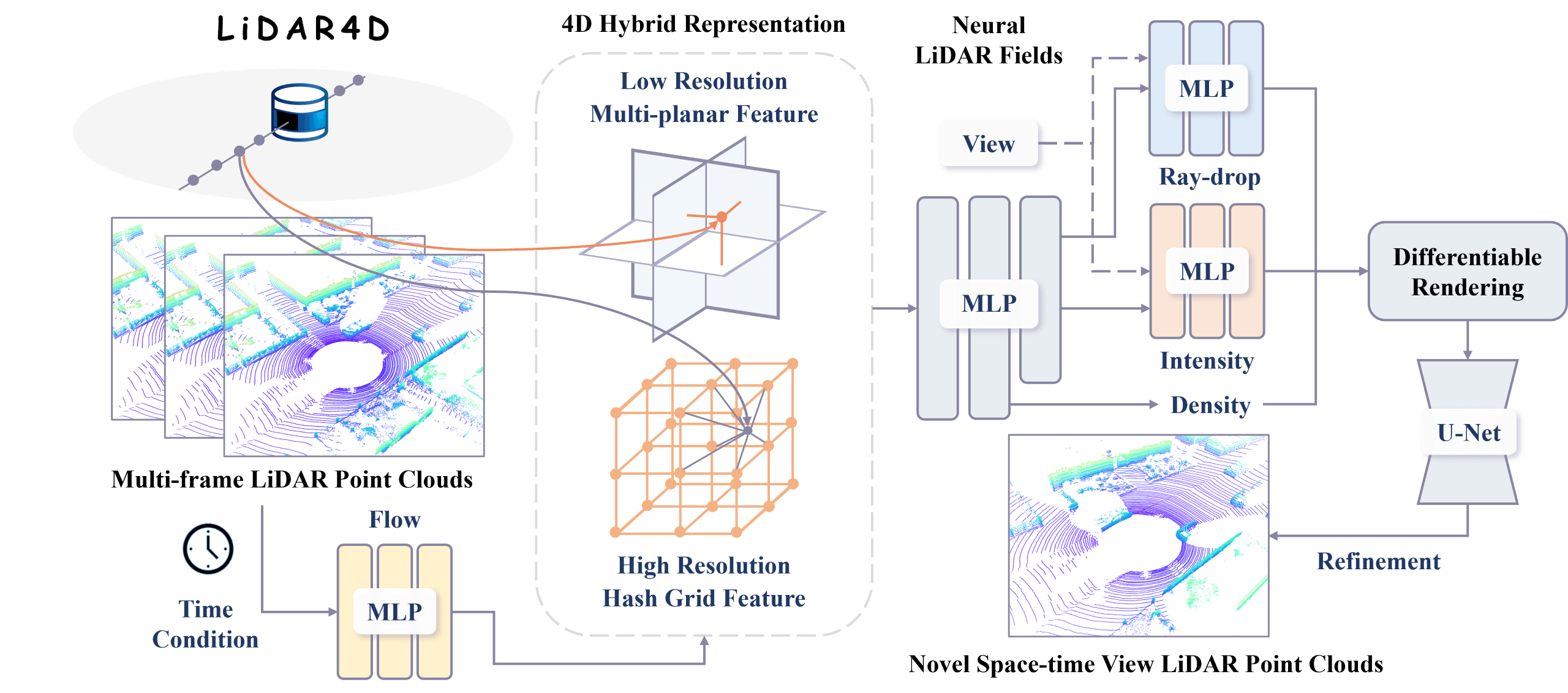}
  \caption{\textbf{Overview of our proposed LiDAR4D.} For large-scale autonomous driving scenarios, we utilize the 4D hybrid representation, which combines low-resolution multi-planar features and high-resolution hash grid features to achieve effective reconstruction. Then, multi-level spatio-temporal features aggregated by flow MLP are fed into neural LiDAR fields for density, intensity and ray-drop probability prediction. Finally, novel space-time view LiDAR point clouds are synthesized via differentiable rendering. Furthermore, we construct geometric constraints derived from point clouds for temporal consistency and the global optimization of ray-drop for generation realism.}
  \label{fig:overview}
\vspace{-.1cm}
\end{figure*}

\textbf{LiDAR Simulation}. Traditional simulators~\cite{dosovitskiy2017carla,shah2018airsim,koenig2004gazebo} such as CARLA are based on physics engines, which can generate LiDAR point clouds via ray casting within handcrafted virtual environments. However, it has diversity limitations and a heavy reliance on costly 3D assets. And there is still a large domain gap compared to real-world data. Thus, several recent works~\cite{manivasagam2020lidarsim,li2023pcgen,guillard2022learning} further narrowed this gap by reconstructing the scene from real data before simulation. LiDARsim~\cite{manivasagam2020lidarsim} reconstructs the mesh surfel representation and employs a neural network to learn the ray-drop characteristics. Besides, it is noted that there are other surface reconstruction works like NKSR~\cite{huang2023nksr} that can convert LiDAR point clouds into mesh representations. Nonetheless, these explicit reconstruction works are troublesome for recovering precise surfaces in large-scale complex scenes, which further leads to a decrease in the accuracy of point cloud synthesis. Instead, PCGen~\cite{li2023pcgen} directly reconstructs from the point clouds, followed by rendering in a rasterization-like manner and first peak averaging. Although it preserves the original information better, the rendering point clouds remain relatively noisy. Moreover, all these explicit methods mentioned above are only applicable to static scenes. In contrast, our approach implicitly reconstructs the continuous representation via space-time neural radiance fields, which achieves higher-quality realistic point cloud synthesis and gets rid of static reconstruction limitations.

\noindent
\textbf{Neural Radiance Fields}. Considerable recent research~\cite{mildenhall2021nerf,barron2021mipnerf,liu2020nsvf,sun2022dvgo,fridovich2022plenoxels,hu2023trimiprf,chan2022eg3d,chen2022tensorf,muller2022instantngp} based on neural radiance fields has led to breakthroughs as well as remarkable achievements in novel view synthesis (NVS) tasks. A wide variety of neural representations based on MLPs~\cite{mildenhall2021nerf,barron2021mipnerf}, voxel grids~\cite{liu2020nsvf,sun2022dvgo,fridovich2022plenoxels}, tri-planes~\cite{hu2023trimiprf,chan2022eg3d}, vector decomposition~\cite{chen2022tensorf}, and multi-level hash grids~\cite{muller2022instantngp} have been fully exploited for reconstruction and synthesis. Yet, most of the work focuses on object-centered reconstruction of small indoor scenes. Subsequently, several works~\cite{barron2022mipnerf360,rematas2022urf,wang2023fegr} gradually extended it to large-scale outdoor scenarios. Despite this, neural radiance fields typically suffer from geometric ambiguity with RGB image inputs. Therefore, DS-NeRF~\cite{deng2022dsnerf} and DDP-NeRF~\cite{roessle2022ddpnerf} introduce the depth prior to enhancing efficiency, and URF~\cite{rematas2022urf} also utilizes LiDAR point clouds to facilitate reconstruction. In this paper, we employ novel hybrid representations and neural LiDAR fields to reconstruct large-scale scenarios for LiDAR NVS.

\noindent
\textbf{NeRF for LiDAR NVS}. Very recently, a few studies~\cite{yang2023unisim,zhang2023nerflidar,tao2023lidarnerf,huang2023nfl} have pioneered in novel view synthesis of LiDAR point clouds based on neural radiance fields, significantly surpassing traditional simulation methods. Among them, NeRF-LiDAR~\cite{zhang2023nerflidar} and UniSim~\cite{yang2023unisim} require both RGB images and LiDAR point clouds as inputs and reconstruct the driving scene with photometric loss and depth supervision. Subsequently, novel-view LiDAR point clouds can be generated through neural depth rendering. Among LiDAR-only methods, LiDAR-NeRF~\cite{tao2023lidarnerf} and NFL~\cite{huang2023nfl} firstly proposed the differentiable LiDAR NVS framework, which reconstructed depth, intensity, and ray-drop probability simultaneously. Nevertheless, these approaches~\cite{zhang2023nerflidar,tao2023lidarnerf,huang2023nfl} are restricted to static scene reconstruction and incapable of handling dynamic objects such as moving vehicles. Although UniSim~\cite{yang2023unisim} does support dynamic scenes, it is largely limited by the need for ground-truth labeling of 3D object detection and decoupling the background and dynamic objects before reconstruction. Instead, our research focuses on LiDAR-only inputs for dynamic scene reconstruction and novel space-time view synthesis without the help of RGB images or ground-truth labels. And it's noteworthy that NFL~\cite{huang2023nfl} has contributed significantly to the detailed physical modeling of LiDAR, such as beam divergence and secondary returns, which is orthogonal to ours and could be beneficial to all LiDAR NVS works.

\noindent
\textbf{Dynamic Scene Reconstruction}. A substantial amount of research~\cite{pumarola2021dnerf,park2021nerfies,park2021hypernerf,fang2022tineuvox,fridovich2023kplanes,shao2023tensor4d,li2021nsff,li2023dynibar,turki2023suds} has been devoted to expanding neural radiance fields to encompass dynamic scene reconstruction.  In general, dynamic NeRFs can be broadly categorized into two groups. One is the deformable neural radiance fields~\cite{pumarola2021dnerf,park2021nerfies,park2021hypernerf,fang2022tineuvox} that map coordinates into the canonical space via continuous deformation fields. While the decoupling of deformation and radiance fields simplifies the optimization, establishing accurate long-distance correspondence remains challenging. The other is spatio-temporal neural fields~\cite{fridovich2023kplanes,shao2023tensor4d,li2021nsff,li2023dynibar}, which consider time as an additional dimensional input to construct a 4D spatio-temporal representation. Thus, it is flexible to simultaneously model appearance, geometry, and motion as a continuous time-varying function. Most previous work has concentrated on relatively smaller displacements indoors, whereas large-scale vehicle movements in autonomous driving scenarios are even more challenging. Furthermore, our work is also the first to introduce dynamic neural radiance fields into the LiDAR NVS task.

\section{Methodology}
\label{sec: method}
In this section, we start with the problem formulation of novel LiDAR view synthesis and the preliminary for NeRFs. Following this, a detailed description of our proposed LiDAR4D framework is provided. 

\noindent
\textbf{Problem Formulation}. In the dynamic driving scenario, given the collected LiDAR point cloud sequence $S=\{S_{0}, S_{1}, ..., S_{n-1}\}$ ($S_i \in \mathbb{R}^{K\times4}$), along with the corresponding sensor poses $P_s=\{P_{0}, P_{1}, ..., P_{n-1}\}$ ($P_i \in SE(3)$) and timestamps $T_s=\{t_{0}, t_{1}, ..., t_{n-1}\}$ ($t_i \in \mathbb{R}$) as inputs. Each single LiDAR frame $S_i$ contains $K$ points of 3D coordinates $\mathbf{x}$ and 1D reflection intensity $\rho$. 
  
The goal of LiDAR4D is to reconstruct this dynamic scene as a continuous implicit representation based on neural fields. Furthermore, given a novel sensor pose $P_{novel}$ and any moment $t_{novel}$, LiDAR4D performs neural rendering to synthesize the LiDAR point cloud $S_{novel}$ with intensities under the novel space-time view.

\noindent
\textbf{Preliminary for NeRF}. Neural radiance fields, NeRFs for short, take 5D inputs of position $\mathbf{x} \in \mathbb{R}^{3}$ and viewing direction $(\theta, \phi)$ as inputs and establish the mapping to the volume density $\sigma$ and color $\mathbf{c}$. Afterward, it performs volume rendering to estimate pixel values and synthesize images in unknown novel views. In detail, it emits a light ray $\mathbf{r}$ from the sensor center $\mathbf{o}$ with the direction $\mathbf{d}$, \ie,  $\mathbf{r} (t) = \mathbf{o}  + t \mathbf{d}$, and then integrates the neural field outputs of $N$ samples along this ray to approximate the pixel color $\mathcal{C}$. The volume rendering function can be formed as follows:
\begin{align}
    \hat{\mathcal C}(\mathbf{r}) = \sum_{i=1}^NT_i(1-e^{-\sigma_i\delta_i}) \mathbf c_i, \  T_i = \exp ( -\sum_{j=1}^{i-1}\sigma_j\delta_j )
    \label{eq:volume_rendering}
\end{align}
where $T$ indicates the accumulated transmittance, $\sigma$ denotes the density and $\delta$ refers to the distance between samples.

\begin{figure}[t]
\centering
  \includegraphics[width=0.475\textwidth]{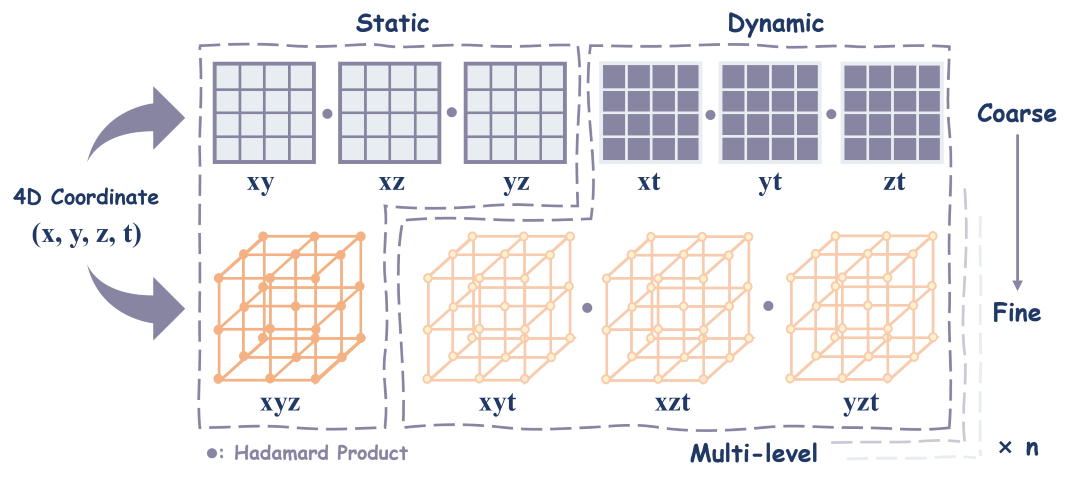}
  \vspace{-.5cm}
  \caption{\textbf{4D decomposition of hybrid planar-grid 
 representation.} Dynamic features can be further aggregated using flow MLP.}
  \label{fig:4d_decompose}
  \vspace{-.3cm}
\end{figure}  

\subsection{LiDAR4D Overview}
\label{sec: overview}

Following neural radiance fields, our proposed LiDAR4D reconstructs the point cloud scene into an implicit continuous representation. Differing from original NeRFs with photometric loss for RGB images, we redefine the neural fields based on LiDAR, which are dubbed neural LiDAR fields. As depicted in \Cref{fig:overview}, it focuses on modeling the geometric depth, reflection intensity, and ray-drop probability of LiDAR point clouds. For large-scale dynamic driving scenarios, LiDAR4D combines coarse-resolution multi-planar features with high-resolution hash grid representation to achieve efficient and effective reconstruction. Then, we lift it to 4D and introduce temporal information encoding for novel space-time view synthesis. To ensure geometry-aware and time-consistent results, we additionally incorporate explicit geometric constraints derived from point clouds. Ultimately, we predict the ray-drop probability for each ray and perform global refinement with a runtime-optimized U-Net to improve generation realism.
  
\begin{figure}[t]
\centering
  \includegraphics[width=0.425\textwidth]{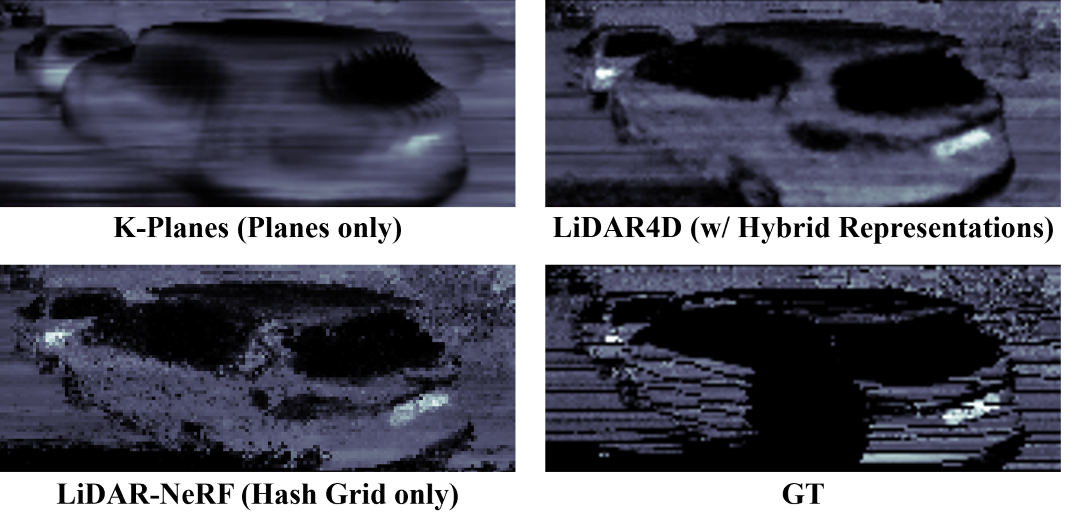}
  \vspace{-.1cm}
  \caption{\textbf{Qualitative comparison for the hybrid representation.} Compared to the noisy intensity reconstruction of LiDAR-NeRF and the blurry one of K-Planes, our hybrid representation achieves more precise and smooth results.}
  \label{fig:hybrid_vis}
  \vspace{-.25cm}
\end{figure}

\subsection{4D Hybrid Planar-Grid Representation}
\label{sec: 4d Hybrid Planar-Grid Representation}

\Cref{fig:4d_decompose} illustrates how our proposed novel hybrid representation breaks down the 4D space into \textit{planar} and \mbox{\textit{hash grid}} features, which further subdivide into \textit{static} and \textit{dynamic} ones. Different from the reconstruction of small indoor objects, large-scale autonomous driving scenes place higher demands on the representation ability and resolution of the features.  However, the dense grid representation such as TiNeuVox~\cite{fang2022tineuvox} is unscalable for large-scale scenarios due to its cubically growing complexity. Therefore, we follow K-planes~\cite{fridovich2023kplanes} and decompose the scene space into a combination of features in multiple orthogonal planes to drastically reduce the parameter quantities. The planar feature can be obtained as follows: 
\begin{equation} \label{planar feature}
    \mathbf{f}_\mathrm{planar} = {\mathcal S}\left(\mathbf{V}, \left(x,y,z,t\right)\right),
    \mathbf{V} \in \mathbb{R}^{(3 M^{2} + 3 M H) C}
\end{equation} 
where $\mathbf{V}$ stores features with $M$ spatial resolution, $H$ temporal resolution and $C$ channels. $\mathcal{S}$ refers to the sampling function that projects 4D coordinates into the corresponding planes (\textit{xy, xz, yz, xt, yt, zt}) and interpolates features bilinearly. Static (\textit{xy, xz, yz}) and dynamic (\textit{xt, yt, zt}) features are multiplied separately by Hadamard product and multiscale features are concatenated in a coarse-to-fine manner.

Nonetheless, for scenes spanning hundreds of meters, this improvement in resolution remains inadequate, especially for the high-frequency intensity reconstruction. Owing to the hash grids proposed in Instant-NGP~\cite{muller2022instantngp}, an explicit grid structure with ultra-high resolution is possible. Additionally, the sparsity of the LiDAR point cloud scene substantially avoids the adverse effects of hash collisions. 
\begin{equation} \label{hash feature}
    \mathbf{f}_\mathrm{hash} = {\mathcal S}\left(\mathbf{G}, \left(x, y, z, t\right)\right), 
    \mathbf{G} \in \mathbb{R}^{(M^{3} + 3 M^{2} H) C}
\end{equation} 
where the dense grid $\mathbf{G}$ will be further compressed into limited storage via hash mapping for parameter reduction. Similarly, the 4D coordinates are projected into static (\textit{xyz}) and dynamic (\textit{xyt, xzt, yzt}) multi-level hash grids before trilinear interpolation and concatenation, where the dynamic features are multiplied using Hadamard product.

However, it's notable that pure hash grid representation still suffers from visual artifacts and noisy reconstruction results (as shown in \Cref{fig:hybrid_vis}), which impede the construction of accurate object geometry. In light of this, we adopt multi-planar features at lower resolutions for overall smooth representation and employ hash grids at higher resolutions to handle finer details, ultimately achieving high accuracy and efficiency in large-scale scene reconstruction.

\subsection{Scene Flow Prior}
\label{sec: scene flow prior}
To enhance the temporal consistency of the current 4D spatio-temporal representations, we further incorporate a flow MLP~\cite{li2021nsfp,zheng2023neuralpci} for motion estimation. It takes the encoded spatio-temporal coordinates as input and constructs the mapping from coordinate fields $\mathbb{R}^{4}$ to motion fields $\mathbb{R}^{3}$.
\begin{equation} \label{flow mlp}
    \Delta \mathbf{x} = \mathrm{MLP}_\mathrm{flow}(\gamma(\mathbf{x}, t)), \quad \mathbf{x}' = \mathbf{x} + \Delta \mathbf{x}
\end{equation} 

Since the vehicle motion range may span a long distance in autonomous driving scenarios, it is extremely hard to establish long-term correspondences to the canonical space in deformable neural radiance fields. Thus, as with Li \etal~\cite{li2021nsff,li2023dynibar}, we utilize the flow MLP to predict motion only between adjacent frames and aggregate multi-frame \textit{dynamic} features to achieve time-consistent reconstruction.

In addition, explicit geometric constraints can be further derived from the input LiDAR point clouds. By feeding point clouds into the flow MLP to produce the scene flow prediction, we can regulate the chamfer distance as a geometric loss ($\mathcal{L}_{\mathrm{flow}}$). It imposes motion prior and additional supervision on LiDAR4D, thus accomplishing the geometry-aware reconstruction. Chamfer Distance between two frames of point cloud ${S}$ and $\hat{S}$ is defined as follows:
\begin{small}
\begin{equation}
\setlength\belowdisplayskip{0.05cm}
\label{eq: CD}
\begin{split}
    \mathrm{CD}=\frac{1}{K} \sum_{\hat{p}_{i} \in \hat{S}} \min _{p_{i} \in S}\left\|\hat{p}_{i}-p_{i}\right\|^{2}_{2}
    +\frac{1}{K} \sum_{p_{i} \in S} \min _{\hat{p}_{i} \in \hat{S}}\left\|p_{i}-\hat{p}_{i}\right\|^{2}_{2}
\end{split}
\end{equation}
\end{small}

\vspace{-0.2cm}

\begin{small}
\begin{equation} \label{flow loss}
  \mathcal{L}_{\mathrm{flow}}=\sum_{j\in\pm1}{\mathrm{CD}(S_{i}+\mathrm{MLP}_\mathrm{flow}(S_{i}),S_{i+j})}, i\in(0,n-1)
\end{equation} 
\end{small}

\subsection{Neural LiDAR Fields}
\label{sec: neural lidar fields}
LiDAR emits laser pulses and measures the time-of-flight (ToF) to determine object distance, along with the intensity of reflected lights. Spinning LiDAR has a 360-degree horizontal field of view (FOV) and a limited range of vertical field of view, which perceives the environment with certain angular resolution lasers. In the same way for neural LiDAR fields, we transmit lasers at specific angular intervals within the FOV, using the center of the LiDAR sensor as the origin $\mathbf{o}$. The direction $\mathbf{d}$ of the laser is determined by the azimuth angle $\theta$ and elevation angle $\phi$ under the polar coordinate system, which is shown below.
\begin{align}
    \mathbf{d} &= (\cos\theta \cos \phi, \ \sin \theta \sin \phi, \ \cos \phi)^T
    \label{eq:lidar_direction}
\end{align}
  
Then we query 3D point coordinates sampled along the laser and feed them into neural fields to predict the density at the corresponding location. Following this, the density along the ray is integrated to obtain the expectation of depth value $\mathcal{D}$, which serves as the laser beam's return distance.
\begin{equation} \label{eq_render_depth}
    \hat{\mathcal{D}}(\mathbf{r}) = \sum_{i=1}^{N}T_i \left( 1 - e^{-\sigma_i\delta_i} \right) z_i, \quad 
    \alpha_i = 1- e^{-\sigma_i\delta_i}
\end{equation} 
where $z_i$ is the depth value of queried points on the ray $\mathbf{r}$, and $\alpha$ is the definition of the opacity.
  
In addition, we predict the intensity $\mathcal{I}$ and ray-drop probability $\mathcal{P}$ separately for each point and similarly conduct alpha-composition along the ray. 
\begin{equation} \label{eq_render_intensity}
    \hat{\mathcal{I}}(\mathbf{r}) = \sum_{i=1}^{N}T_i \alpha_i i_i, \quad 
    \hat{\mathcal{P}}(\mathbf{r}) = \sum_{i=1}^{N}T_i \alpha_i p_i
\end{equation} 
where $i_i$ and $p_i$ are the point-wise intensity and ray-drop probability outputs of LiDAR4D. We use separate MLPs to take temporal aggregated planar and hash features, as well as position-encoded viewpoints as inputs for prediction.

\vspace{-0.1cm}

\begin{equation}
i = \mathrm{MLP}_\mathrm{intensity}(\mathbf{f}_\mathrm{planar}, \mathbf{f}_\mathrm{hash}, \gamma(\mathbf{d}))
\end{equation} 

\vspace{-0.3cm}

\begin{equation}
\rho = \mathrm{MLP}_\mathrm{raydrop}(\mathbf{f}_\mathrm{planar}, \mathbf{f}_\mathrm{hash}, \gamma(\mathbf{d}))
\end{equation} 

\vspace{-0.5cm}

\begin{small}
\begin{equation}
\gamma (x) = (\sin(2^{0}x), \cos(2^{0}x), ..., \sin(2^{L-1}x) ,\cos(2^{L-1}x))
\end{equation} 
\end{small}

\begin{figure}[t]
\centering
  \includegraphics[width=0.475\textwidth]{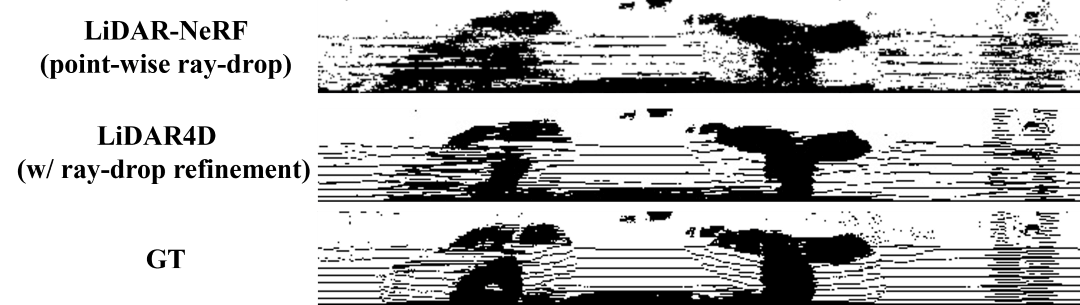}
  \vspace{-.5cm}
  \caption{\textbf{Qualitative comparison for the ray-drop refinement.} The point-wise prediction of ray-drop probability by MLP cannot preserve global patterns well. Instead, LiDAR4D drastically improves generation realism via runtime-optimized U-Net. }
  \label{fig:raydrop_vis}
  \vspace{-.2cm}
\end{figure}

\definecolor{best_result}{rgb}{0.96, 0.57, 0.58}
\definecolor{second_result}{rgb}{0.98, 0.78, 0.57}
\definecolor{third_result}{rgb}{1.0, 1.0, 0.56}

\begin{table*}[th]
\centering
\resizebox{\textwidth}{!}{
\renewcommand{\arraystretch}{1.25}
\begin{tabular}{ccccccccccccccc}
\hline \multirow[c]{2}{*}{ Method } & \multirow[c]{2}{*}{ Type } & \multicolumn{2}{c}{ Point Cloud } & \multicolumn{5}{c}{ Depth } & \multicolumn{5}{c}{ Intensity } \\
&    & CD$\downarrow$ & F-score$\uparrow$ & RMSE$\downarrow$ & MedAE$\downarrow$ & LPIPS$\downarrow$ & SSIM$\uparrow$ & PSNR$\uparrow$ & RMSE$\downarrow$ & MedAE$\downarrow$ & LPIPS$\downarrow$ & SSIM$\uparrow$ & PSNR$\uparrow$ \\

\hline LiDARsim~\cite{manivasagam2020lidarsim} & $\mathcal{E.}$ / $\mathcal{S.}$ / $\mathcal{M.}$ & 3.2228 & 0.7157 & 6.9153 & 0.1279 & 0.2926 & 0.6342 & 21.4608 & 0.1666 & 0.0569 & 0.3276 & 0.3502 & 15.5853 \\

NKSR~\cite{huang2023nksr} & $\mathcal{E.}$ / $\mathcal{S.}$ / $\mathcal{M.}$ & 1.8982 & 0.6855 & 5.8403 & 0.0996 & \cellcolor{second_result}0.2752 & 0.6409 & 23.0368 & 0.1742 & 0.0590 & 0.3337 & 0.3517 & 15.2081 \\

PCGen~\cite{li2023pcgen} & $\mathcal{E.}$ / $\mathcal{S.}$ & 0.4636 & 0.8023 & 5.6583 & 0.2040 & 0.5391 & 0.4903 & 23.1675 & 0.1970 & 0.0763 & 0.5926 & 0.1351 & 14.1181 \\

LiDAR-NeRF~\cite{tao2023lidarnerf} & $\mathcal{I.}$ / $\mathcal{S.}$ & 0.1438 & 0.9091 & 4.1753 & 0.0566 & 0.2797 & 0.6568 & 25.9878 & 0.1404 & 0.0443 & \cellcolor{second_result}0.3135 & \cellcolor{second_result}0.3831 & 17.1549 \\

D-NeRF~\cite{pumarola2021dnerf} & $\mathcal{I.}$ / $\mathcal{D.}$  & 0.1442 & \cellcolor{second_result}0.9128 & \cellcolor{second_result}4.0194 & 0.0508 & 0.3061 & \cellcolor{second_result}0.6634 & \cellcolor{second_result}26.2344 & 0.1369 & \cellcolor{second_result}0.0440 & 0.3409 & 0.3748 & \cellcolor{second_result}17.3554 \\

TiNeuVox-B~\cite{fang2022tineuvox} & $\mathcal{I.}$ / $\mathcal{D.}$ & 0.1748 & 0.9059 & 4.1284 & \cellcolor{second_result}0.0502 & 0.3427 & 0.6514 & 26.0267 & \cellcolor{second_result}0.1363 & 0.0453 & 0.4365 & 0.3457 & 17.3535 \\

K-Planes~\cite{fridovich2023kplanes} & $\mathcal{I.}$ / $\mathcal{D.}$ &  \cellcolor{second_result}0.1302 & 0.9123 & 4.1322 & 0.0539 & 0.3457 & 0.6385 & 26.0236 & 0.1415 & 0.0498 & 0.4081 & 0.3008 & 17.0167 \\

\textbf{LiDAR4D (Ours)} & $\mathcal{I.}$ / $\mathcal{D.}$ & \cellcolor{best_result}\textbf{0.1089} & \cellcolor{best_result}\textbf{0.9272} & \cellcolor{best_result}\textbf{3.5256} & \cellcolor{best_result}\textbf{0.0404} & \cellcolor{best_result}\textbf{0.1051} & \cellcolor{best_result}\textbf{0.7647} & \cellcolor{best_result}\textbf{27.4767} & \cellcolor{best_result}\textbf{0.1195} & \cellcolor{best_result}\textbf{0.0327} & \cellcolor{best_result}\textbf{0.1845} & \cellcolor{best_result}\textbf{0.5304} & \cellcolor{best_result}\textbf{18.5561}  \\

\hline
\end{tabular}}
\vspace{-0.1cm}
\caption{\textbf{Quantitative comparison on KITTI-360 dataset}. We compare our method to different types of previous approaches and color the top results as \colorbox{best_result}{best} and \colorbox{second_result}{second best}. $\mathcal{E}$: Explicit, $\mathcal{I}$ : Implicit,  $\mathcal{S}$: Static,  $\mathcal{D}$: Dynamic, $\mathcal{M}$: Mesh. }  
\label{exp:kitti360_dynamic}
\end{table*}

\definecolor{best_result}{rgb}{0.96, 0.57, 0.58}
\definecolor{second_result}{rgb}{0.98, 0.78, 0.57}
\definecolor{third_result}{rgb}{1.0, 1.0, 0.56}

\begin{table*}[ht]
\centering
\resizebox{\textwidth}{!}{
\renewcommand{\arraystretch}{1.25}
\begin{tabular}{ccccccccccccccc}
\hline \multirow[c]{2}{*}{ Method } & \multirow[c]{2}{*}{ Type } & \multicolumn{2}{c}{ Point Cloud } & \multicolumn{5}{c}{ Depth } & \multicolumn{5}{c}{ Intensity } \\
&    & CD$\downarrow$ & F-score$\uparrow$ & RMSE$\downarrow$ & MedAE$\downarrow$ & LPIPS$\downarrow$ & SSIM$\uparrow$ & PSNR$\uparrow$ & RMSE$\downarrow$ & MedAE$\downarrow$ & LPIPS$\downarrow$ & SSIM$\uparrow$ & PSNR$\uparrow$ \\

\hline LiDARsim~\cite{manivasagam2020lidarsim} & $\mathcal{E.}$ / $\mathcal{S.}$ / $\mathcal{M.}$ & 12.1383 & 0.6512 & 10.5539 & 0.3572 & 0.1871 & 0.5653 & 17.7841 & 0.0659 & 0.0115 & 0.1160 & 0.5170 & 23.7791 \\

NKSR~\cite{huang2023nksr} & $\mathcal{E.}$ / $\mathcal{S.}$ / $\mathcal{M.}$ & 11.4910 & 0.6178 & 9.3731 & 0.5763 & 0.2111 & 0.5637 & 18.7774 & 0.0680 & 0.0119 & 0.1290 & 0.5031 & 23.4905 \\

PCGen~\cite{li2023pcgen} & $\mathcal{E.}$ / $\mathcal{S.}$ & 2.1998 & 0.6341 & 8.8364 & 0.4011 & 0.1792 & 0.5440 & 19.2799 & 0.0768 & 0.0147 & 0.1308 & 0.4410 & 22.4428 \\

LiDAR-NeRF~\cite{tao2023lidarnerf} & $\mathcal{I.}$ / $\mathcal{S.}$ & 0.3225 & 0.8576 & 7.1566 & 0.0338 & \cellcolor{second_result}0.0702 & 0.7188 & 21.2129 & 0.0467 & \cellcolor{second_result}0.0076 & \cellcolor{second_result}0.0483 & 0.7264 & 26.9927 \\

D-NeRF~\cite{pumarola2021dnerf} & $\mathcal{I.}$ / $\mathcal{D.}$  & 0.3296 & 0.8513 & 7.1089 & 0.0368 & 0.0789 & 0.7130 & 21.2594 & 0.0467 & 0.0080 & 0.0492 & 0.7180 & 26.9951 \\

TiNeuVox-B~\cite{fang2022tineuvox} & $\mathcal{I.}$ / $\mathcal{D.}$ & 0.3920 & 0.8627 & 7.2093 & 0.0290 & 0.1549 & 0.6873 & 21.0932 & 0.0462 & 0.0080 & 0.1294 & 0.7107 & 26.8620 \\

K-Planes~\cite{fridovich2023kplanes} & $\mathcal{I.}$ / $\mathcal{D.}$ & \cellcolor{second_result}0.2982 & \cellcolor{second_result}0.8887 & \cellcolor{second_result}6.7960 & \cellcolor{best_result}\textbf{0.0209} & 0.1218 & \cellcolor{second_result}0.7258 & \cellcolor{second_result}21.6203 & \cellcolor{second_result}0.0438 & \cellcolor{second_result}0.0076 & 0.1127 & \cellcolor{second_result}0.7364 & \cellcolor{second_result}27.4227 \\

\textbf{LiDAR4D (Ours)} & $\mathcal{I.}$ / $\mathcal{D.}$ & \cellcolor{best_result}\textbf{0.2443} & \cellcolor{best_result}\textbf{0.8915} & \cellcolor{best_result}\textbf{6.7831} & \cellcolor{second_result}0.0258 & \cellcolor{best_result}\textbf{0.0569} & \cellcolor{best_result}\textbf{0.7396} & \cellcolor{best_result}\textbf{21.7189} & \cellcolor{best_result}\textbf{0.0426} & \cellcolor{best_result}\textbf{0.0071} & \cellcolor{best_result}\textbf{0.0459} & \cellcolor{best_result}\textbf{0.7498} & \cellcolor{best_result}\textbf{27.7977} \\
\hline
\end{tabular}}
\vspace{-0.1cm}
\caption{\textbf{Quantitative comparison on NuScenes dataset}. The notations are consistent with the KITTI-360~\Cref{exp:kitti360_dynamic} above.} 
\vspace{-0.4cm}
\label{exp:nuscenes}
\end{table*}

\subsection{Ray-drop Refinement}
\label{sec: ray-drop refinement}
During laser ranging, a portion of the emitted rays is not reflected back to the sensor, which is termed the ray-drop characteristic. In fact, the ray-drop of LiDAR is significantly impacted by various aspects, including distance, surface properties and sensor noise. As in LiDAR-NeRF~\cite{tao2023lidarnerf}, ray-drop prediction is directly accomplished with a point-wise MLP head, which is essentially noisy and unreliable. To address this issue, we employ the U-Net~\cite{ronneberger2015unet} with residuals to refine the ray-drop mask globally and better preserve the consistent pattern across regions. It takes the \textit{full} ray-drop probability, depth and intensity prediction of LiDAR4D as inputs (different from previous work) and refines the final mask via binary cross-entropy loss as follows:
\begin{equation} \label{raydrop refinement loss}
    \mathcal{L}_\mathrm{refine} = \mathrm{BCELoss}\left(\hat{\mathcal{M}}_\mathrm{Pred}, \mathcal{M}_\mathrm{GT}\right)
\end{equation} 
where $\mathcal{M}$ indicates the global mask rendered from range view and $\mathcal{M}_{GT}$ is calculated from the input point clouds.

We emphasize that the lightweight network is randomly initialized and optimized at runtime efficiently for reconstruction. As illustrated in \Cref{fig:raydrop_vis}, the global optimization greatly improves the prediction results and further enhances the fidelity of the generated LiDAR point cloud.

\subsection{Optimization}
\label{sec: optimization}

For the optimization of LiDAR4D, the total reconstruction loss is the weighted combination of the depth loss, intensity loss, ray-drop loss, flow loss and refinement loss, which can be formalized as follows:

\begin{equation} \label{depth loss}
  \mathcal{L}_{\mathrm{depth}} = \sum_{\mathbf{r}\in R}\begin{Vmatrix} \hat{D}(\mathbf{r}) - D(\mathbf{r}) \end{Vmatrix}_1
\end{equation} 

\vspace{-0.1cm}

\begin{equation} \label{intensity loss}
  \mathcal{L}_{\mathrm{intensity}} = \sum_{\mathbf{r}\in R}\begin{Vmatrix} \hat{I}(\mathbf{r}) - I(\mathbf{r}) \end{Vmatrix}_2^2
\end{equation} 

\vspace{-0.1cm}

\begin{equation} \label{raydrop loss}
  \mathcal{L}_{\mathrm{raydrop}} = \sum_{\mathbf{r}\in R}\begin{Vmatrix} \hat{P}(\mathbf{r}) - P(\mathbf{r}) \end{Vmatrix}_2^2
\end{equation} 

\vspace{-0.2cm}

\begin{equation}
\begin{split}
\label{optimization loss}
    \mathcal{L} & = \lambda_{\alpha}\mathcal{L}_\mathrm{depth} + \lambda_{\beta}\mathcal{L}_\mathrm{intensity} + \lambda_{\gamma}\mathcal{L}_\mathrm{raydrop} \\ & + \lambda_{\eta}\mathcal{L}_\mathrm{flow} + 
    \lambda_\mathrm{r}\mathcal{L}_\mathrm{refine} 
\end{split}
\end{equation}

\begin{figure}[ht]
\centering
  \includegraphics[width=0.475\textwidth]{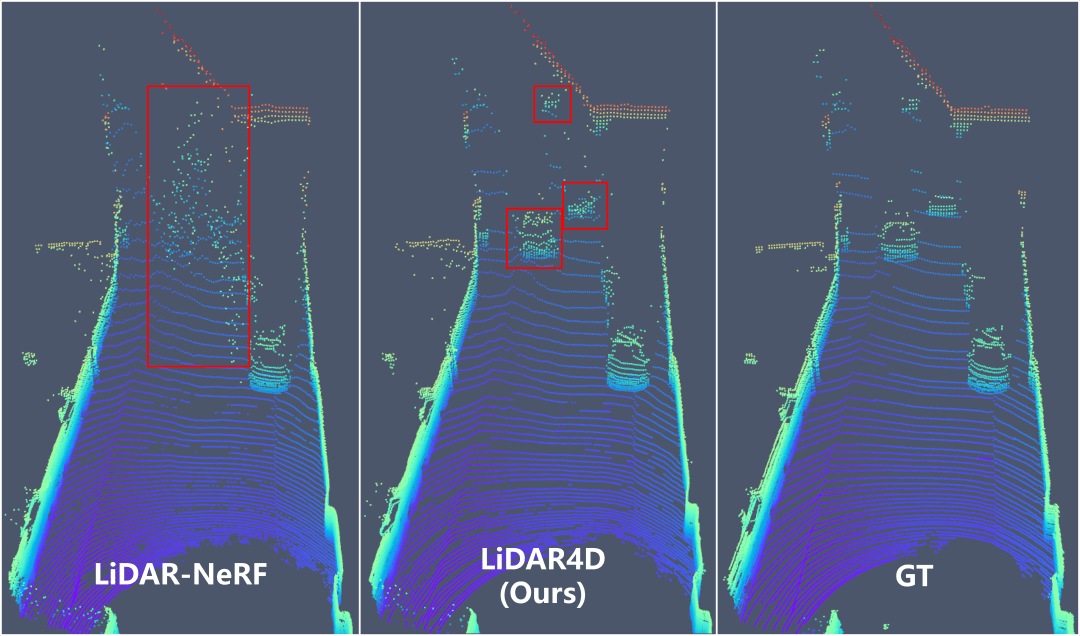}
  \vspace{-.5cm}
  \caption{\textbf{Qualitative novel view LiDAR point cloud synthesis results on KITTI-360 dataset.} As highlighted in the red bounding box, LiDAR-NeRF fails to reconstruct the dynamic vehicles. In contrast, LiDAR4D generates more accurate geometry for moving cars, even in sparse point clouds far away.}
  \label{fig:pointcloud_vis}
\vspace{-.1cm}
\end{figure}

\begin{figure*}[th]
\centering
  \includegraphics[width=0.9\textwidth,height=0.35\textheight]{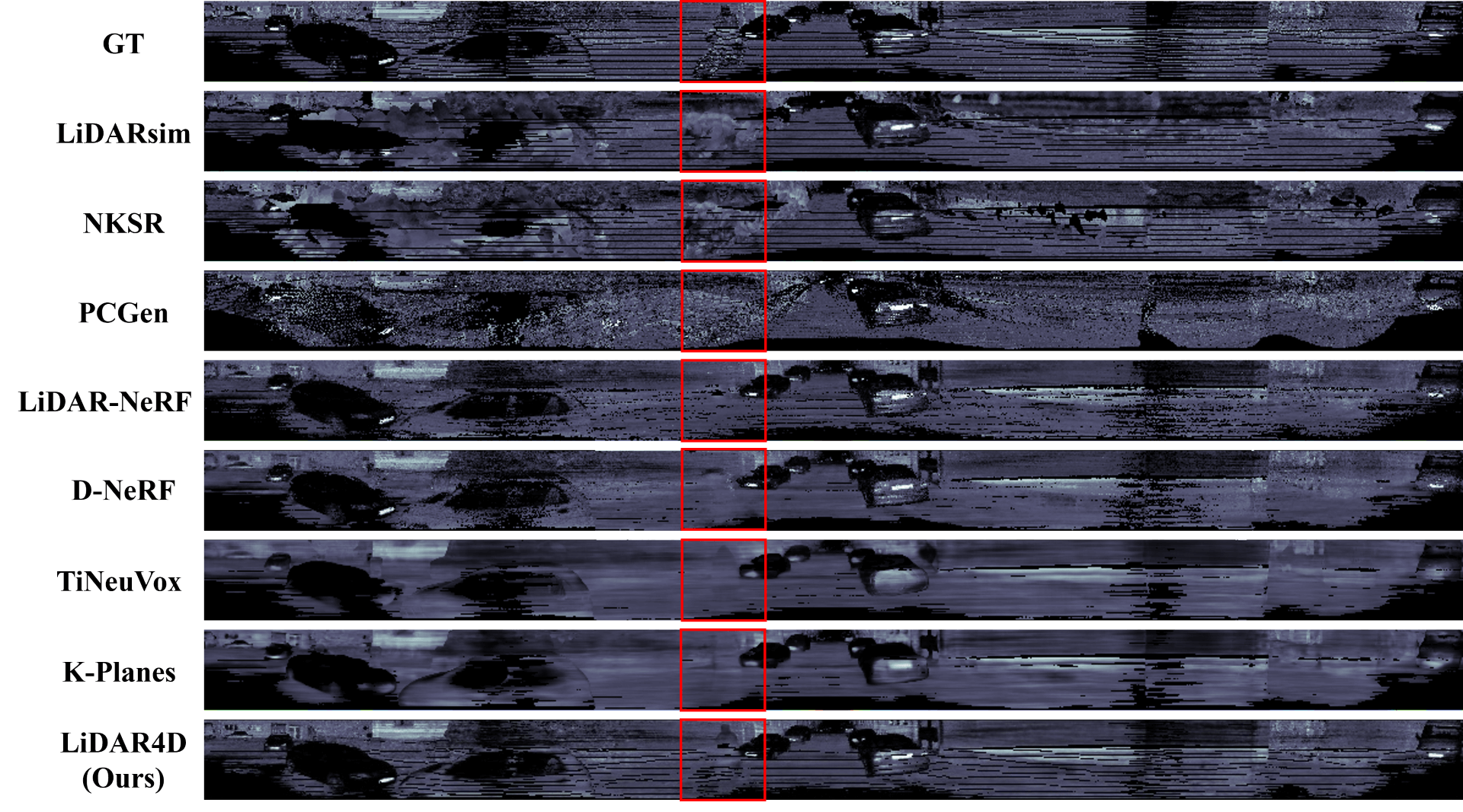}
  \caption{\textbf{Qualitative comparison for LiDAR intensity reconstruction and synthesis.}  }
  \label{fig:IntensityRecon}
  \vspace{0.2cm}
\end{figure*}

\begin{figure*}[th]
\centering
  \includegraphics[width=0.9\textwidth,height=0.35\textheight]{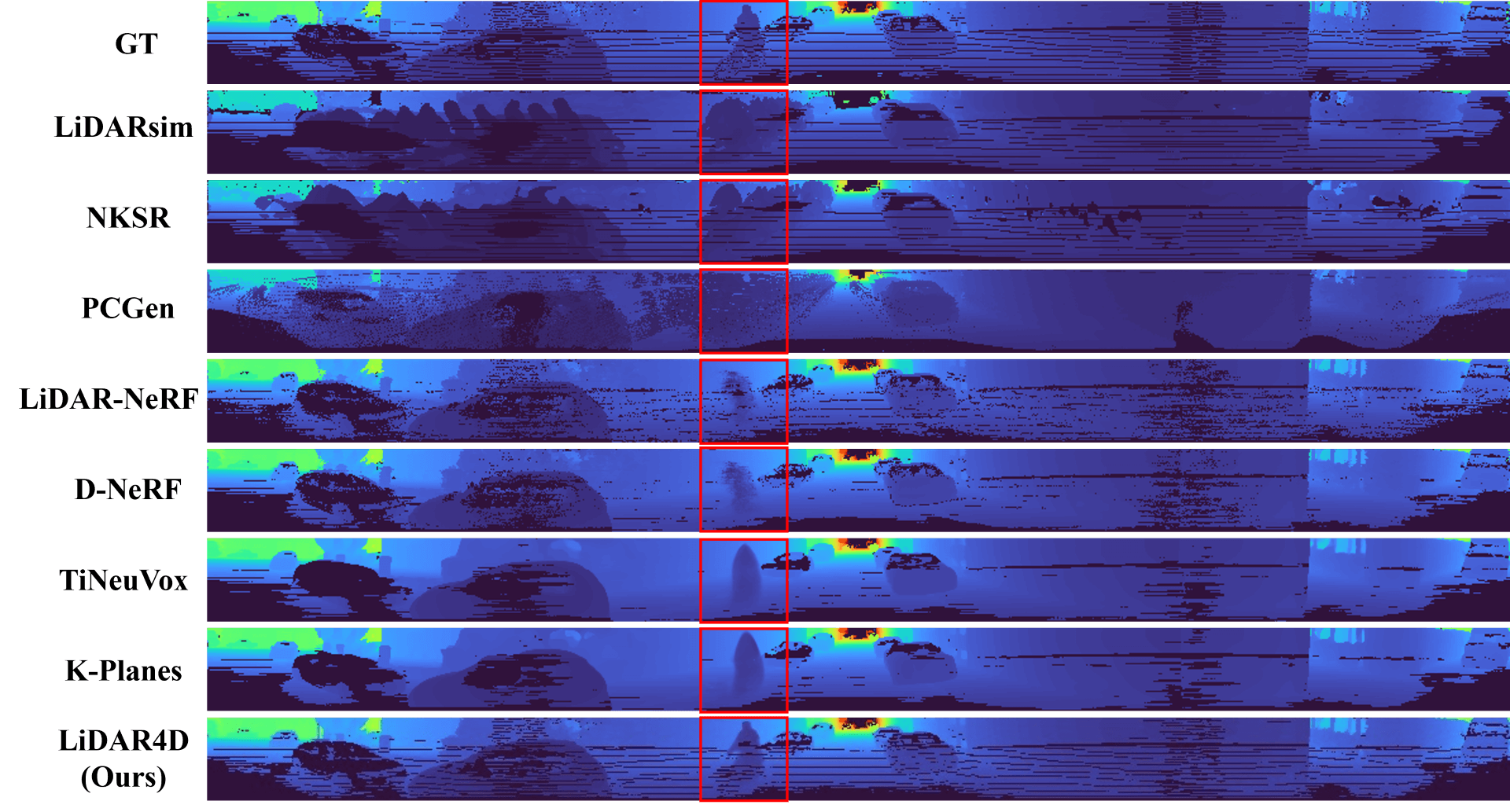}
  \caption{\textbf{Qualitative comparison for LiDAR depth reconstruction and synthesis.}  }
  \label{fig:DepthRecon}
  \vspace{0.2cm}
\end{figure*}  


\begin{table*}[ht]
\centering
\resizebox{\textwidth}{!}{
\renewcommand{\arraystretch}{1.3}
\begin{tabular}{ccccccccccccccc}
\hline \multirow[c]{2}{*}{ Method } & \multirow[c]{2}{*}{ Type } & \multicolumn{2}{c}{ Point Cloud } & \multicolumn{5}{c}{ Depth } & \multicolumn{5}{c}{ Intensity } \\
&    & CD$\downarrow$ & F-score$\uparrow$ & RMSE$\downarrow$ & MedAE$\downarrow$ & LPIPS$\downarrow$ & SSIM$\uparrow$ & PSNR$\uparrow$ & RMSE$\downarrow$ & MedAE$\downarrow$ & LPIPS$\downarrow$ & SSIM$\uparrow$ & PSNR$\uparrow$ \\

\hline LiDARsim~\cite{manivasagam2020lidarsim} & $\mathcal{E.}$ / $\mathcal{S.}$ / $\mathcal{M.}$ & 2.2249 & 0.8667 & 6.5470 & 0.0759 & 0.2289 & \underline{0.7157} & 21.7746 & 0.1532 & 0.0506 & 0.2502 & \underline{0.4479} & \underline{16.3045} \\

NKSR~\cite{huang2023nksr} & $\mathcal{E.}$ / $\mathcal{S.}$ / $\mathcal{M.}$ & 0.5780 & 0.8685 & 4.6647 & 0.0698 & 0.2295 & 0.7052 & 22.5390 & 0.1565 & \underline{0.0536} & \underline{0.2429} & 0.4200 & 16.1159 \\

PCGen~\cite{li2023pcgen} & $\mathcal{E.}$ / $\mathcal{S.}$ & 0.2090 & 0.8597 & 4.8838 & 0.1785 & 0.5210 & 0.5062 & 24.3050 & 0.2005 & 0.0818 & 0.6100 & 0.1248 & 13.9606 \\

LiDAR-NeRF~\cite{tao2023lidarnerf} & $\mathcal{I.}$ / $\mathcal{S.}$  & \underline{0.0923} & \underline{0.9226} & \underline{3.6801} & \underline{0.0667} & \underline{0.3523} & 0.6043 & \underline{26.7663} & \underline{0.1557} & 0.0549 & 0.4212 & 0.2768 & 16.1683 \\

\textbf{LiDAR4D (Ours)} & $\mathcal{I.}$ / $\mathcal{D.}$ &  \textbf{0.0894} &  \textbf{0.9264} &  \textbf{3.2370} &  \textbf{0.0507} &  \textbf{0.1313} &  \textbf{0.7218} &  \textbf{27.8840} &  \textbf{0.1343} &  \textbf{0.0404} &  \textbf{0.2127} &  \textbf{0.4698} &  \textbf{17.4529}  \\

\hline
\end{tabular}}
\caption{\textbf{Quantitative comparison on KITTI-360 \textit{Static} Scene Sequence}. We bold the best results and underline the second-best.} 
\label{exp:kitti360_static}
\end{table*}

\section{Experiments}
\label{sec: experiments}
\subsection{Experimental Setup}
\label{sec: Experimental Setup}
\textbf{Datasets}. We conducted comprehensive experiments on the public autonomous driving datasets KITTI-360~\cite{liao2022kitti} and NuScenes~\cite{caesar2020nuscenes}, from which we collected multiple dynamic point cloud sequences containing largely moving vehicles. KITTI-360 is equipped with a LIDAR of 64-beam, a 26.4-degree vertical FOV, and an acquisition frequency of 10Hz. We selected 51 consecutive frames as a single scene and held out 4 samples at a 10-frame interval for NVS evaluation. Meanwhile, Nuscenes' LiDAR has 32 beams, a 40-degree vertical FOV, and a 20-Hz acquisition frequency. To cover a larger range of reconstruction, i.e., spanning 100$\sim$200 meters, we still selected samples at a frequency of 10 Hz, which is consistent with the KITTI-360.  

\noindent
\textbf{Baselines}. We present a comprehensive comparison of LiDAR4D with different types of baselines, encompassing explicit and implicit reconstruction approaches as well as dynamic NeRFs. We reproduce the mesh-based reconstruction method LiDARsim~\cite{manivasagam2020lidarsim} and also replace the surface reconstruction model with the state-of-the-art method NKSR~\cite{huang2023nksr} for convincing. It's noted that we employ U-Net trained from original point clouds to further predict ray-drop for these two methods. Meanwhile, PCGen~\cite{li2023pcgen} reconstructs directly based on the point clouds and predicts ray-drop with the MLP. LiDAR-NeRF~\cite{tao2023lidarnerf} is our primary comparison, and we directly adopt the official implements. In addition, we migrate dynamic neural radiance field methods such as D-NeRF~\cite{pumarola2021dnerf}, K-Planes~\cite{fridovich2023kplanes}, and TiNeuVox~\cite{fang2022tineuvox} to LiDAR NVS pipeline for a thorough comparison.

\noindent
\textbf{Metrics}. We offer multi-faceted metrics for evaluation. Chamfer Distance~\cite{fan2017chamfer} measures the 3D geometric error between the generated and the ground-truth point clouds by nearest neighbor, and we also report the F-score value with an error threshold of 5cm. In addition, we introduce Root Mean Square Error (RMSE) and Median Absolute Error (MedAE) to calculate the pixel-by-pixel error of the projected range images, as well as LPIPS~\cite{zhang2018lpips}, SSIM~\cite{wang2004ssim} and PSNR to measure the overall variance. We evaluate both the depth and intensity reconstruction results.

\subsection{Implementation Details}

Consistent with LiDAR-NeRF~\cite{tao2023lidarnerf}, the entire point cloud scene is scaled within the unit cube space. And we uniformly sampled 768 points along each laser. The optimization of LiDAR4D is implemented on Pytorch~\cite{paszke2019pytorch} with Adam~\cite{kingma2014adam} optimizer. The maximum iteration is set to 30k for each scene, with a batch size of 1024 rays, followed by the fast ray-drop refinement with 300 epochs.
We construct multi-planar representations following K-Planes~\cite{fridovich2023kplanes} and hash grids built on tiny-cuda-nn~\cite{tiny-cuda-nn}. The multi-level features of planes and hash grids are concatenated before feeding into MLPs. All experiments were conducted on a single NVIDIA GeForce RTX 4090 GPU. For more implementation details, please refer to \textit{Supplementary Material}.

\subsection{Evaluation of Novel-View LiDAR Synthesis}
\textbf{Results on KITTI-360 dataset}. The quantitative comparison on KITTI-360 dataset is displayed in \Cref{exp:kitti360_dynamic}. Our proposed LiDAR4D exhibits remarkable performance across all metrics in comparison to prior SOTA methods, demonstrating its superiority in dynamic reconstruction. In comparison to LiDAR-NeRF, our approach has led to a 24.3\% reduction in the CD error of the novel-view point cloud synthesis. As illustrated in~\Cref{fig:pointcloud_vis}, LiDAR4D achieves accurate reconstruction of every dynamic vehicle, whereas LiDAR-NeRF encounters failure. As shown in \Cref{fig:IntensityRecon,fig:DepthRecon}, the resolution limitation causes blurring and oversmooth results of dynamic NeRFs such as TiNeuVox and K-Planes. This demonstrates again the effectiveness of our designed hybrid representation for large-scale scenes. Additionally, we follow the setting in LiDAR-NeRF to repeat experiments in static scenarios, and \Cref{exp:kitti360_static} verifies that there is no performance degradation in LiDAR4D.

\noindent
\textbf{Results on NuScenes dataset}. To further validate the generalizability of LiDAR4D, we conducted the same experiment on NuScenes. As illustrated in \Cref{exp:nuscenes}, our method still achieves the best reconstruction quality even with a completely different LiDAR configuration. Ultimately, our method still excels in the reconstruction of depth and intensity, as evidenced by the 24.2\% reduction in CD error to LiDAR-NeRF. Please refer to \textit{Supplementary Material} for more quantitative and qualitative experimental results on KITTI-360 and NuScenes datasets, as well as the ablation study and further applications.

\vspace{-0.1cm}

\section{Limitations}
\label{sec: limitations}
Despite the fact that LiDAR4D has exhibited exceptional performance in a substantial number of experiments, the long-distance vehicle motion and occlusion problem of point clouds remain open questions. There is still a significant gap in the reconstruction of dynamic objects compared to static ones. In addition, foreground and background may be difficult to separate well. Furthermore, based on real-world datasets, \textit{quantitative} evaluation of NVS is limited within the ego-car trajectory and does not allow for the decoupling of novel spatial and temporal view synthesis.

\section{Conclusion}
\label{sec: conclusion}
In this paper, we revisit the limitations of existing LiDAR NVS methods and propose a novel framework to address three major challenges, namely dynamic reconstruction, large-scale scene characterization, and realistic synthesis. Our proposed method LiDAR4D proves its superiority under extensive experiments, achieving geometry-aware and time-consistent reconstruction of large-scale dynamic point cloud scenes and generating novel space-time view LiDAR point clouds closer to the real distribution. We believe that more future work will focus on combining LiDAR point clouds with neural radiance fields and explore more possibilities for dynamic scene reconstruction and synthesis.

\vspace{0.15cm}

\noindent\textbf{Acknowledgment.} This work is supported by the National Natural Science Foundation of China (No.62372329), in part by the National Key Research and Development Program of China (No.2021YFB2501104), in part by Shanghai Rising Star Program (No.21QC1400900), in part by Tongji-Qomolo Autonomous Driving Commercial Vehicle Joint Lab Project, and in part by Xiaomi Young Talents \mbox{Program}.

{
    \small
    \bibliographystyle{ieeenat_fullname}
    \bibliography{LiDAR4D}

\begin{thebibliography}{46}
\providecommand{\natexlab}[1]{#1}
\providecommand{\url}[1]{\texttt{#1}}
\expandafter\ifx\csname urlstyle\endcsname\relax
  \providecommand{\doi}[1]{doi: #1}\else
  \providecommand{\doi}{doi: \begingroup \urlstyle{rm}\Url}\fi

\bibitem[Barron et~al.(2021)Barron, Mildenhall, Tancik, Hedman, Martin-Brualla, and Srinivasan]{barron2021mipnerf}
Jonathan~T Barron, Ben Mildenhall, Matthew Tancik, Peter Hedman, Ricardo Martin-Brualla, and Pratul~P Srinivasan.
\newblock Mip-nerf: A multiscale representation for anti-aliasing neural radiance fields.
\newblock In \emph{Proceedings of the IEEE/CVF International Conference on Computer Vision}, pages 5855--5864, 2021.

\bibitem[Barron et~al.(2022)Barron, Mildenhall, Verbin, Srinivasan, and Hedman]{barron2022mipnerf360}
Jonathan~T Barron, Ben Mildenhall, Dor Verbin, Pratul~P Srinivasan, and Peter Hedman.
\newblock Mip-nerf 360: Unbounded anti-aliased neural radiance fields.
\newblock In \emph{Proceedings of the IEEE/CVF Conference on Computer Vision and Pattern Recognition}, pages 5470--5479, 2022.

\bibitem[Caesar et~al.(2020)Caesar, Bankiti, Lang, Vora, Liong, Xu, Krishnan, Pan, Baldan, and Beijbom]{caesar2020nuscenes}
Holger Caesar, Varun Bankiti, Alex~H Lang, Sourabh Vora, Venice~Erin Liong, Qiang Xu, Anush Krishnan, Yu Pan, Giancarlo Baldan, and Oscar Beijbom.
\newblock nuscenes: A multimodal dataset for autonomous driving.
\newblock In \emph{Proceedings of the IEEE/CVF conference on computer vision and pattern recognition}, pages 11621--11631, 2020.

\bibitem[Chan et~al.(2022)Chan, Lin, Chan, Nagano, Pan, De~Mello, Gallo, Guibas, Tremblay, Khamis, et~al.]{chan2022eg3d}
Eric~R Chan, Connor~Z Lin, Matthew~A Chan, Koki Nagano, Boxiao Pan, Shalini De~Mello, Orazio Gallo, Leonidas~J Guibas, Jonathan Tremblay, Sameh Khamis, et~al.
\newblock Efficient geometry-aware 3d generative adversarial networks.
\newblock In \emph{Proceedings of the IEEE/CVF Conference on Computer Vision and Pattern Recognition}, pages 16123--16133, 2022.

\bibitem[Chen et~al.(2022)Chen, Xu, Geiger, Yu, and Su]{chen2022tensorf}
Anpei Chen, Zexiang Xu, Andreas Geiger, Jingyi Yu, and Hao Su.
\newblock Tensorf: Tensorial radiance fields.
\newblock In \emph{European Conference on Computer Vision}, pages 333--350. Springer, 2022.

\bibitem[Deng et~al.(2022)Deng, Liu, Zhu, and Ramanan]{deng2022dsnerf}
Kangle Deng, Andrew Liu, Jun-Yan Zhu, and Deva Ramanan.
\newblock Depth-supervised nerf: Fewer views and faster training for free.
\newblock In \emph{Proceedings of the IEEE/CVF Conference on Computer Vision and Pattern Recognition}, pages 12882--12891, 2022.

\bibitem[Dosovitskiy et~al.(2017)Dosovitskiy, Ros, Codevilla, Lopez, and Koltun]{dosovitskiy2017carla}
Alexey Dosovitskiy, German Ros, Felipe Codevilla, Antonio Lopez, and Vladlen Koltun.
\newblock Carla: An open urban driving simulator.
\newblock In \emph{Conference on robot learning}, pages 1--16. PMLR, 2017.

\bibitem[Fan et~al.(2017)Fan, Su, and Guibas]{fan2017chamfer}
Haoqiang Fan, Hao Su, and Leonidas~J Guibas.
\newblock A point set generation network for 3d object reconstruction from a single image.
\newblock In \emph{Proceedings of the IEEE conference on computer vision and pattern recognition}, pages 605--613, 2017.

\bibitem[Fang et~al.(2022)Fang, Yi, Wang, Xie, Zhang, Liu, Nie{\ss}ner, and Tian]{fang2022tineuvox}
Jiemin Fang, Taoran Yi, Xinggang Wang, Lingxi Xie, Xiaopeng Zhang, Wenyu Liu, Matthias Nie{\ss}ner, and Qi Tian.
\newblock Fast dynamic radiance fields with time-aware neural voxels.
\newblock In \emph{SIGGRAPH Asia 2022 Conference Papers}, pages 1--9, 2022.

\bibitem[Fischler and Bolles(1981)]{fischler1981ransac}
Martin~A Fischler and Robert~C Bolles.
\newblock Random sample consensus: a paradigm for model fitting with applications to image analysis and automated cartography.
\newblock \emph{Communications of the ACM}, 24\penalty0 (6):\penalty0 381--395, 1981.

\bibitem[Fridovich-Keil et~al.(2022)Fridovich-Keil, Yu, Tancik, Chen, Recht, and Kanazawa]{fridovich2022plenoxels}
Sara Fridovich-Keil, Alex Yu, Matthew Tancik, Qinhong Chen, Benjamin Recht, and Angjoo Kanazawa.
\newblock Plenoxels: Radiance fields without neural networks.
\newblock In \emph{Proceedings of the IEEE/CVF Conference on Computer Vision and Pattern Recognition}, pages 5501--5510, 2022.

\bibitem[Fridovich-Keil et~al.(2023)Fridovich-Keil, Meanti, Warburg, Recht, and Kanazawa]{fridovich2023kplanes}
Sara Fridovich-Keil, Giacomo Meanti, Frederik~Rahb{\ae}k Warburg, Benjamin Recht, and Angjoo Kanazawa.
\newblock K-planes: Explicit radiance fields in space, time, and appearance.
\newblock In \emph{Proceedings of the IEEE/CVF Conference on Computer Vision and Pattern Recognition}, pages 12479--12488, 2023.

\bibitem[Guillard et~al.(2022)Guillard, Vemprala, Gupta, Miksik, Vineet, Fua, and Kapoor]{guillard2022learning}
Beno{\^\i}t Guillard, Sai Vemprala, Jayesh~K Gupta, Ondrej Miksik, Vibhav Vineet, Pascal Fua, and Ashish Kapoor.
\newblock Learning to simulate realistic lidars.
\newblock In \emph{2022 IEEE/RSJ International Conference on Intelligent Robots and Systems (IROS)}, pages 8173--8180. IEEE, 2022.

\bibitem[Hu et~al.(2023)Hu, Wang, Ma, Yang, Gao, Liu, and Ma]{hu2023trimiprf}
Wenbo Hu, Yuling Wang, Lin Ma, Bangbang Yang, Lin Gao, Xiao Liu, and Yuewen Ma.
\newblock Tri-miprf: Tri-mip representation for efficient anti-aliasing neural radiance fields.
\newblock In \emph{Proceedings of the IEEE/CVF International Conference on Computer Vision}, pages 19774--19783, 2023.

\bibitem[Huang et~al.(2023{\natexlab{a}})Huang, Gojcic, Atzmon, Litany, Fidler, and Williams]{huang2023nksr}
Jiahui Huang, Zan Gojcic, Matan Atzmon, Or Litany, Sanja Fidler, and Francis Williams.
\newblock Neural kernel surface reconstruction.
\newblock In \emph{Proceedings of the IEEE/CVF Conference on Computer Vision and Pattern Recognition}, pages 4369--4379, 2023{\natexlab{a}}.

\bibitem[Huang et~al.(2023{\natexlab{b}})Huang, Gojcic, Wang, Williams, Kasten, Fidler, Schindler, and Litany]{huang2023nfl}
Shengyu Huang, Zan Gojcic, Zian Wang, Francis Williams, Yoni Kasten, Sanja Fidler, Konrad Schindler, and Or Litany.
\newblock Neural lidar fields for novel view synthesis.
\newblock \emph{arXiv preprint arXiv:2305.01643}, 2023{\natexlab{b}}.

\bibitem[Kingma and Ba(2014)]{kingma2014adam}
Diederik~P Kingma and Jimmy Ba.
\newblock Adam: A method for stochastic optimization.
\newblock \emph{arXiv preprint arXiv:1412.6980}, 2014.

\bibitem[Koenig and Howard(2004)]{koenig2004gazebo}
Nathan Koenig and Andrew Howard.
\newblock Design and use paradigms for gazebo, an open-source multi-robot simulator.
\newblock In \emph{2004 IEEE/RSJ international conference on intelligent robots and systems (IROS)(IEEE Cat. No. 04CH37566)}, pages 2149--2154. IEEE, 2004.

\bibitem[Li et~al.(2023{\natexlab{a}})Li, Ren, and Liu]{li2023pcgen}
Chenqi Li, Yuan Ren, and Bingbing Liu.
\newblock Pcgen: Point cloud generator for lidar simulation.
\newblock In \emph{2023 IEEE International Conference on Robotics and Automation (ICRA)}, pages 11676--11682. IEEE, 2023{\natexlab{a}}.

\bibitem[Li et~al.(2021{\natexlab{a}})Li, Kaesemodel~Pontes, and Lucey]{li2021nsfp}
Xueqian Li, Jhony Kaesemodel~Pontes, and Simon Lucey.
\newblock Neural scene flow prior.
\newblock \emph{Advances in Neural Information Processing Systems}, 34:\penalty0 7838--7851, 2021{\natexlab{a}}.

\bibitem[Li et~al.(2021{\natexlab{b}})Li, Niklaus, Snavely, and Wang]{li2021nsff}
Zhengqi Li, Simon Niklaus, Noah Snavely, and Oliver Wang.
\newblock Neural scene flow fields for space-time view synthesis of dynamic scenes.
\newblock In \emph{Proceedings of the IEEE/CVF Conference on Computer Vision and Pattern Recognition}, pages 6498--6508, 2021{\natexlab{b}}.

\bibitem[Li et~al.(2023{\natexlab{b}})Li, Wang, Cole, Tucker, and Snavely]{li2023dynibar}
Zhengqi Li, Qianqian Wang, Forrester Cole, Richard Tucker, and Noah Snavely.
\newblock Dynibar: Neural dynamic image-based rendering.
\newblock In \emph{Proceedings of the IEEE/CVF Conference on Computer Vision and Pattern Recognition}, pages 4273--4284, 2023{\natexlab{b}}.

\bibitem[Liao et~al.(2022)Liao, Xie, and Geiger]{liao2022kitti}
Yiyi Liao, Jun Xie, and Andreas Geiger.
\newblock Kitti-360: A novel dataset and benchmarks for urban scene understanding in 2d and 3d.
\newblock \emph{IEEE Transactions on Pattern Analysis and Machine Intelligence}, 45\penalty0 (3):\penalty0 3292--3310, 2022.

\bibitem[Liu et~al.(2020)Liu, Gu, Zaw~Lin, Chua, and Theobalt]{liu2020nsvf}
Lingjie Liu, Jiatao Gu, Kyaw Zaw~Lin, Tat-Seng Chua, and Christian Theobalt.
\newblock Neural sparse voxel fields.
\newblock \emph{Advances in Neural Information Processing Systems}, 33:\penalty0 15651--15663, 2020.

\bibitem[Manivasagam et~al.(2020)Manivasagam, Wang, Wong, Zeng, Sazanovich, Tan, Yang, Ma, and Urtasun]{manivasagam2020lidarsim}
Sivabalan Manivasagam, Shenlong Wang, Kelvin Wong, Wenyuan Zeng, Mikita Sazanovich, Shuhan Tan, Bin Yang, Wei-Chiu Ma, and Raquel Urtasun.
\newblock Lidarsim: Realistic lidar simulation by leveraging the real world.
\newblock In \emph{Proceedings of the IEEE/CVF Conference on Computer Vision and Pattern Recognition}, pages 11167--11176, 2020.

\bibitem[Mildenhall et~al.(2021)Mildenhall, Srinivasan, Tancik, Barron, Ramamoorthi, and Ng]{mildenhall2021nerf}
Ben Mildenhall, Pratul~P Srinivasan, Matthew Tancik, Jonathan~T Barron, Ravi Ramamoorthi, and Ren Ng.
\newblock Nerf: Representing scenes as neural radiance fields for view synthesis.
\newblock \emph{Communications of the ACM}, 65\penalty0 (1):\penalty0 99--106, 2021.

\bibitem[M\"uller(2021)]{tiny-cuda-nn}
Thomas M\"uller.
\newblock {tiny-cuda-nn}, 2021.

\bibitem[M{\"u}ller et~al.(2022)M{\"u}ller, Evans, Schied, and Keller]{muller2022instantngp}
Thomas M{\"u}ller, Alex Evans, Christoph Schied, and Alexander Keller.
\newblock Instant neural graphics primitives with a multiresolution hash encoding.
\newblock \emph{ACM Transactions on Graphics (ToG)}, 41\penalty0 (4):\penalty0 1--15, 2022.

\bibitem[Park et~al.(2021{\natexlab{a}})Park, Sinha, Barron, Bouaziz, Goldman, Seitz, and Martin-Brualla]{park2021nerfies}
Keunhong Park, Utkarsh Sinha, Jonathan~T Barron, Sofien Bouaziz, Dan~B Goldman, Steven~M Seitz, and Ricardo Martin-Brualla.
\newblock Nerfies: Deformable neural radiance fields.
\newblock In \emph{Proceedings of the IEEE/CVF International Conference on Computer Vision}, pages 5865--5874, 2021{\natexlab{a}}.

\bibitem[Park et~al.(2021{\natexlab{b}})Park, Sinha, Hedman, Barron, Bouaziz, Goldman, Martin-Brualla, and Seitz]{park2021hypernerf}
Keunhong Park, Utkarsh Sinha, Peter Hedman, Jonathan~T Barron, Sofien Bouaziz, Dan~B Goldman, Ricardo Martin-Brualla, and Steven~M Seitz.
\newblock Hypernerf: A higher-dimensional representation for topologically varying neural radiance fields.
\newblock \emph{arXiv preprint arXiv:2106.13228}, 2021{\natexlab{b}}.

\bibitem[Paszke et~al.(2019)Paszke, Gross, Massa, Lerer, Bradbury, Chanan, Killeen, Lin, Gimelshein, Antiga, et~al.]{paszke2019pytorch}
Adam Paszke, Sam Gross, Francisco Massa, Adam Lerer, James Bradbury, Gregory Chanan, Trevor Killeen, Zeming Lin, Natalia Gimelshein, Luca Antiga, et~al.
\newblock Pytorch: An imperative style, high-performance deep learning library.
\newblock \emph{Advances in neural information processing systems}, 32, 2019.

\bibitem[Pumarola et~al.(2021)Pumarola, Corona, Pons-Moll, and Moreno-Noguer]{pumarola2021dnerf}
Albert Pumarola, Enric Corona, Gerard Pons-Moll, and Francesc Moreno-Noguer.
\newblock D-nerf: Neural radiance fields for dynamic scenes.
\newblock In \emph{Proceedings of the IEEE/CVF Conference on Computer Vision and Pattern Recognition}, pages 10318--10327, 2021.

\bibitem[Rematas et~al.(2022)Rematas, Liu, Srinivasan, Barron, Tagliasacchi, Funkhouser, and Ferrari]{rematas2022urf}
Konstantinos Rematas, Andrew Liu, Pratul~P Srinivasan, Jonathan~T Barron, Andrea Tagliasacchi, Thomas Funkhouser, and Vittorio Ferrari.
\newblock Urban radiance fields.
\newblock In \emph{Proceedings of the IEEE/CVF Conference on Computer Vision and Pattern Recognition}, pages 12932--12942, 2022.

\bibitem[Roessle et~al.(2022)Roessle, Barron, Mildenhall, Srinivasan, and Nie{\ss}ner]{roessle2022ddpnerf}
Barbara Roessle, Jonathan~T Barron, Ben Mildenhall, Pratul~P Srinivasan, and Matthias Nie{\ss}ner.
\newblock Dense depth priors for neural radiance fields from sparse input views.
\newblock In \emph{Proceedings of the IEEE/CVF Conference on Computer Vision and Pattern Recognition}, pages 12892--12901, 2022.

\bibitem[Ronneberger et~al.(2015)Ronneberger, Fischer, and Brox]{ronneberger2015unet}
Olaf Ronneberger, Philipp Fischer, and Thomas Brox.
\newblock U-net: Convolutional networks for biomedical image segmentation.
\newblock In \emph{Medical Image Computing and Computer-Assisted Intervention--MICCAI 2015: 18th International Conference, Munich, Germany, October 5-9, 2015, Proceedings, Part III 18}, pages 234--241. Springer, 2015.

\bibitem[Shah et~al.(2018)Shah, Dey, Lovett, and Kapoor]{shah2018airsim}
Shital Shah, Debadeepta Dey, Chris Lovett, and Ashish Kapoor.
\newblock Airsim: High-fidelity visual and physical simulation for autonomous vehicles.
\newblock In \emph{Field and Service Robotics: Results of the 11th International Conference}, pages 621--635. Springer, 2018.

\bibitem[Shao et~al.(2023)Shao, Zheng, Tu, Liu, Zhang, and Liu]{shao2023tensor4d}
Ruizhi Shao, Zerong Zheng, Hanzhang Tu, Boning Liu, Hongwen Zhang, and Yebin Liu.
\newblock Tensor4d: Efficient neural 4d decomposition for high-fidelity dynamic reconstruction and rendering.
\newblock In \emph{Proceedings of the IEEE/CVF Conference on Computer Vision and Pattern Recognition}, pages 16632--16642, 2023.

\bibitem[Sun et~al.(2022)Sun, Sun, and Chen]{sun2022dvgo}
Cheng Sun, Min Sun, and Hwann-Tzong Chen.
\newblock Direct voxel grid optimization: Super-fast convergence for radiance fields reconstruction.
\newblock In \emph{Proceedings of the IEEE/CVF Conference on Computer Vision and Pattern Recognition}, pages 5459--5469, 2022.

\bibitem[Tao et~al.(2023)Tao, Gao, Wang, Chen, Hao, Liang, Salzmann, and Yu]{tao2023lidarnerf}
Tang Tao, Longfei Gao, Guangrun Wang, Peng Chen, Dayang Hao, Xiaodan Liang, Mathieu Salzmann, and Kaicheng Yu.
\newblock Lidar-nerf: Novel lidar view synthesis via neural radiance fields.
\newblock \emph{arXiv preprint arXiv:2304.10406}, 2023.

\bibitem[Turki et~al.(2023)Turki, Zhang, Ferroni, and Ramanan]{turki2023suds}
Haithem Turki, Jason~Y Zhang, Francesco Ferroni, and Deva Ramanan.
\newblock Suds: Scalable urban dynamic scenes.
\newblock In \emph{Proceedings of the IEEE/CVF Conference on Computer Vision and Pattern Recognition}, pages 12375--12385, 2023.

\bibitem[Wang et~al.(2004)Wang, Bovik, Sheikh, and Simoncelli]{wang2004ssim}
Zhou Wang, Alan~C Bovik, Hamid~R Sheikh, and Eero~P Simoncelli.
\newblock Image quality assessment: from error visibility to structural similarity.
\newblock \emph{IEEE transactions on image processing}, 13\penalty0 (4):\penalty0 600--612, 2004.

\bibitem[Wang et~al.(2023)Wang, Shen, Gao, Huang, Munkberg, Hasselgren, Gojcic, Chen, and Fidler]{wang2023fegr}
Zian Wang, Tianchang Shen, Jun Gao, Shengyu Huang, Jacob Munkberg, Jon Hasselgren, Zan Gojcic, Wenzheng Chen, and Sanja Fidler.
\newblock Neural fields meet explicit geometric representations for inverse rendering of urban scenes.
\newblock In \emph{Proceedings of the IEEE/CVF Conference on Computer Vision and Pattern Recognition}, pages 8370--8380, 2023.

\bibitem[Yang et~al.(2023)Yang, Chen, Wang, Manivasagam, Ma, Yang, and Urtasun]{yang2023unisim}
Ze Yang, Yun Chen, Jingkang Wang, Sivabalan Manivasagam, Wei-Chiu Ma, Anqi~Joyce Yang, and Raquel Urtasun.
\newblock Unisim: A neural closed-loop sensor simulator.
\newblock In \emph{Proceedings of the IEEE/CVF Conference on Computer Vision and Pattern Recognition}, pages 1389--1399, 2023.

\bibitem[Zhang et~al.(2023)Zhang, Zhang, Kuang, and Zhang]{zhang2023nerflidar}
Junge Zhang, Feihu Zhang, Shaochen Kuang, and Li Zhang.
\newblock Nerf-lidar: Generating realistic lidar point clouds with neural radiance fields.
\newblock \emph{arXiv preprint arXiv:2304.14811}, 2023.

\bibitem[Zhang et~al.(2018)Zhang, Isola, Efros, Shechtman, and Wang]{zhang2018lpips}
Richard Zhang, Phillip Isola, Alexei~A Efros, Eli Shechtman, and Oliver Wang.
\newblock The unreasonable effectiveness of deep features as a perceptual metric.
\newblock In \emph{Proceedings of the IEEE conference on computer vision and pattern recognition}, pages 586--595, 2018.

\bibitem[Zheng et~al.(2023)Zheng, Wu, Lu, Lu, Chen, and Jiang]{zheng2023neuralpci}
Zehan Zheng, Danni Wu, Ruisi Lu, Fan Lu, Guang Chen, and Changjun Jiang.
\newblock Neuralpci: Spatio-temporal neural field for 3d point cloud multi-frame non-linear interpolation.
\newblock In \emph{Proceedings of the IEEE/CVF Conference on Computer Vision and Pattern Recognition}, pages 909--918, 2023.

\end{thebibliography}
}

\clearpage
\setcounter{page}{1}

\appendix

\noindent{\Large \textbf{Appendix}}
\vspace{0.6cm}

\renewcommand{\appendixname}{\appendixname~\Alph{section}}
\renewcommand*{\thefigure}{S\arabic{figure}}
\renewcommand*{\thetable}{S\arabic{table}}

In this document, we start with the \textbf{ablation study} in \Cref{sec:ablation_study} to demonstrate the effectiveness of the proposed key modules as a complement. Following this, we conduct a more comprehensive \textbf{analysis of the qualitative and quantitative experiments} based on \Cref{sec: experiments} and provide additional experimental results in \Cref{sec:additional_exp}. Specific \textbf{implementation details} and dataset information are subsequently presented in \Cref{sec:implementation_details} for reproduction. Finally, we showcase further \textbf{applications} of LiDAR4D in \Cref{sec:applications}, thereby highlighting its versatility, flexibility, and great potential.

\vspace{0.1cm}

\section{Ablation Study}
\label{sec:ablation_study}

The advantages of our method in comparison to LiDAR-NeRF~\cite{tao2023lidarnerf} are illustrated in \Cref{fig:pointcloud_vis,fig:hybrid_vis,fig:raydrop_vis}, which corresponds to the key modules of LiDAR4D, \ie, dynamic reconstruction, hybrid representation, and ray-drop refinement. In order to provide a more rigorous demonstration of the efficacy of our method, we perform the ablation study for each module and present quantitative results in \Cref{exp:kitti360_ablation}.

\begin{figure}[ht]
\centering
  \includegraphics[width=0.475\textwidth]{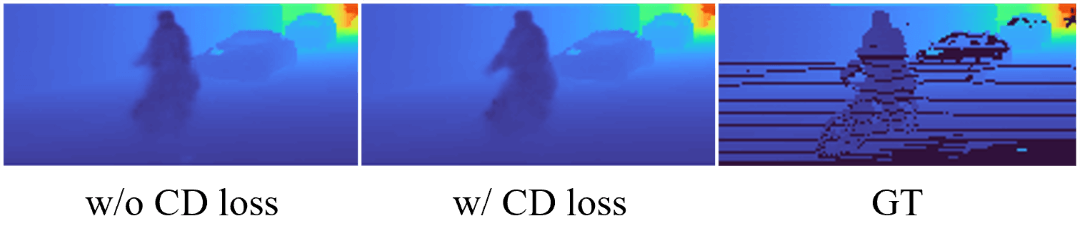}
  \vspace{-.6cm}
  \caption{\textbf{Qualitative comparison for the geometric regularization of CD loss.} }
  \label{fig:CD_loss}
\end{figure}  


\begin{table*}[th]
\centering
\resizebox{\textwidth}{!}{
\renewcommand{\arraystretch}{1.3}
\begin{tabular}{cccccccccccccccc}
\hline \multirow[c]{2}{*}{$\mathcal{H.}$} & \multirow[c]{2}{*}{$\mathcal{D_{T}.}$} & \multirow[c]{2}{*}{$\mathcal{D_{F}.}$} & \multirow[c]{2}{*}{ $\mathcal{R.}$ } & \multicolumn{2}{c}{ Point Cloud } & \multicolumn{5}{c}{ Depth } & \multicolumn{5}{c}{ Intensity } \\
&  &  &  & CD$\downarrow$ & F-score$\uparrow$ & RMSE$\downarrow$ & MedAE$\downarrow$ & LPIPS$\downarrow$ & SSIM$\uparrow$ & PSNR$\uparrow$ & RMSE$\downarrow$ & MedAE$\downarrow$ & LPIPS$\downarrow$ & SSIM$\uparrow$ & PSNR$\uparrow$ \\
\hline

 \ding{55} & \ding{55} & \ding{55} & \ding{55} & 0.1840 & 0.8979 & 4.0602 & 0.0639 & 0.2692 & 0.6483 & 26.1957 & 0.1398 & 0.0431 & 0.2969 & 0.3829 & 17.2018 \\
 
 \checkmark & \ding{55} & \ding{55} & \ding{55} & 0.1429 & 0.9116 & 3.9702 & 0.0499 & 0.2586 & 0.6645 & 26.3647 & 0.1368 & 0.0411 & 0.2760 & 0.4036 & 17.3675 \\

 \checkmark & \checkmark & \ding{55} & \ding{55} & 0.1213 & 0.9221 & 3.6947 & 0.0448 & 0.2397 & 0.7027 & 27.0285 & 0.1286 & 0.0368 & 0.2688 & 0.4553 & 17.8999 \\

\checkmark & \checkmark & \checkmark & \ding{55} & 0.1187 & 0.9260 & 3.6745 & 0.0425 & 0.2130 & 0.7104 & 27.1009 & 0.1281 & 0.0359 & 0.2426 & 0.4726 & 17.9394 \\

 \checkmark & \checkmark & \checkmark & \checkmark & \textbf{0.1089} & \textbf{0.9272} & \textbf{3.5256} & \textbf{0.0404} & \textbf{0.1051} & \textbf{0.7647} & \textbf{27.4767} & \textbf{0.1195} & \textbf{0.0327} & \textbf{0.1845} & \textbf{0.5304} & \textbf{18.5561}  \\

\hline
\end{tabular}}
\caption{\textbf{Ablation study on KITTI-360 Dataset}. $\mathcal{H}$: hybrid representation, $\mathcal{D_T}$: time-conditioned dynamic-part representations, $\mathcal{D_F}$: flow-constrained temporal feature aggregation, $\mathcal{R}$: global ray-drop refinement.} 
\vspace{0.1cm}
\label{exp:kitti360_ablation}
\end{table*}

The results in the first row represent the basic version of LiDAR4D with only hash grid representation, which is similar to LiDAR-NeRF. The introduction of the hybrid representation ($\mathcal{H.}$) significantly enhances the reconstruction quality, especially for the point cloud and depth metrics in \textit{Row 2}. Subsequently, we further adopted time-conditioned dynamic-part representations ($\mathcal{D_T.}$) and flow-constrained temporal feature aggregation ($\mathcal{D_F.}$), which notably strengthened the capability of dynamic reconstruction in \textit{Row 3\&4}. Among them, the incorporation of CD loss as geometric regularization benefits the optimization of flow MLP and leads to more accurate results for dynamic objects, as shown in \Cref{fig:CD_loss}. Ultimately, the global optimization of ray-drop ($\mathcal{R.}$) based on U-Net assists LiDAR4D in achieving SOTA performance in the last row.

\section{Additional Analysis and Experiments}
\label{sec:additional_exp}

\subsection{Quantitative and Qualitative Comparison} 

\begin{figure}[th]
\centering
  \includegraphics[width=0.475\textwidth]{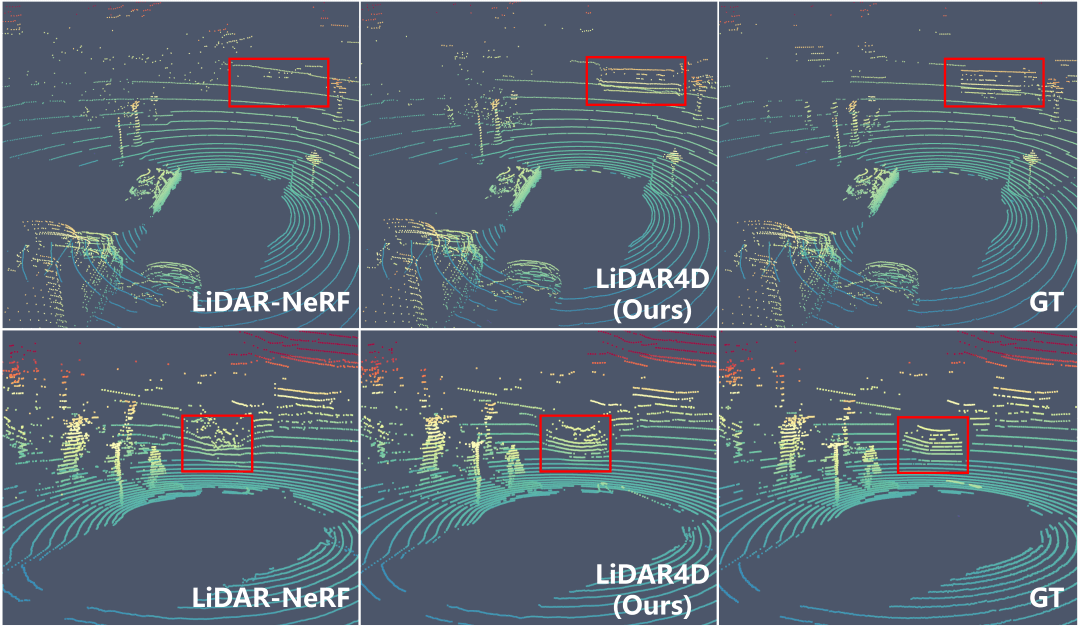}
  \caption{\textbf{Qualitative novel view LiDAR point cloud synthesis results on NuScenes dataset.}}
  \label{fig:pointcloud_nuscenes}
\end{figure} 

\begin{figure*}[th]
\centering
  \includegraphics[width=0.9\textwidth]{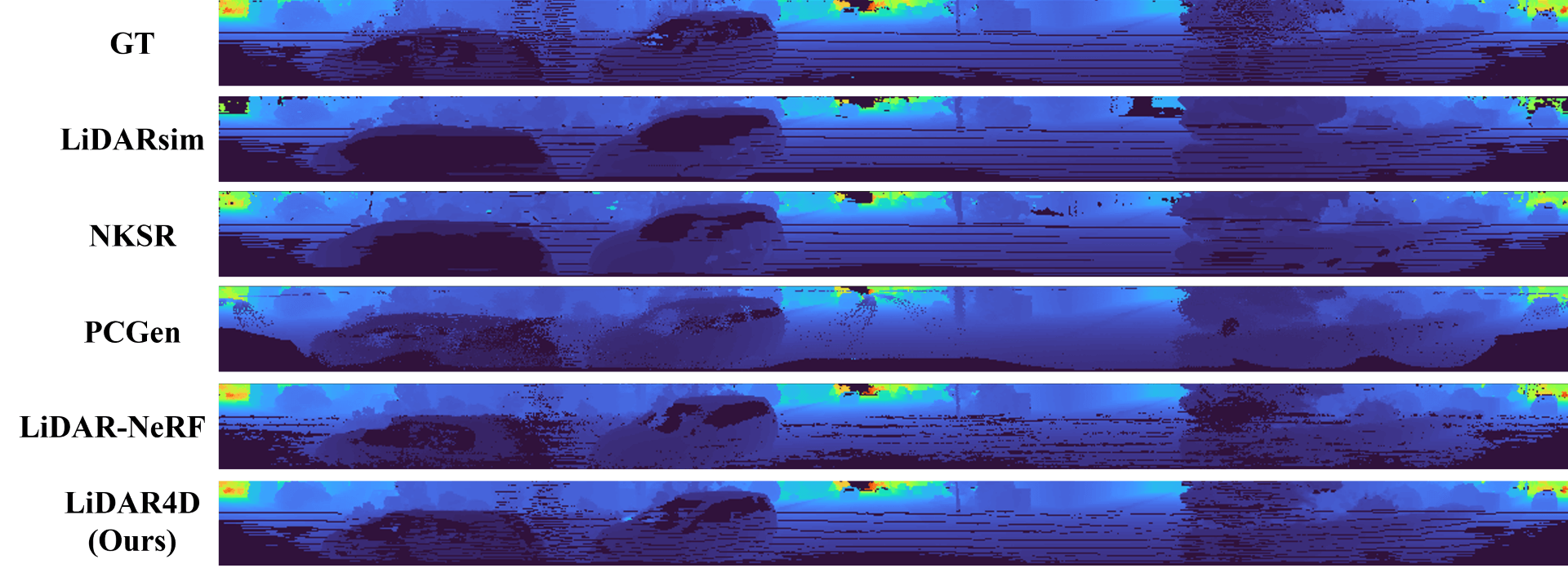}
  \caption{\textbf{Qualitative comparison on KITTI-360 \textit{Static} Scene Sequences.} }
  \label{fig:results_static}
\end{figure*}

\noindent
\textbf{Static Scenes.} As shown in~\Cref{fig:results_static}, traditional explicit reconstruction methods such as LiDARsim~\cite{manivasagam2020lidarsim} convert point cloud scenes into mesh representations but struggle to accurately reconstruct object details in large-scale scenes (\textit{Row 2}). We additionally adopt the state-of-the-art surface reconstruction algorithm NKSR~\cite{huang2023nksr} upon LiDARsim to improve the reconstruction quality. Nevertheless, the novel-view results are still significantly different from the ground truth (\textit{Row 3}). Furthermore, it is unable to establish the correlation between intensity and viewpoint. PC-Gen~\cite{li2023pcgen} reconstructs directly based on the point cloud, while the generated results are heavily affected by noise (\textit{Row 4}). On the contrary, the implicit reconstruction method like LiDAR-NeRF~\cite{tao2023lidarnerf} (\textit{Row 5}) alleviates the challenges above and achieves a substantial lead. Our LiDAR4D further surpasses the previous approaches, especially in reconstruction details such as vehicle shape and window reflections (\textit{Row 6}). The quantitative results illustrated in~\Cref{exp:kitti360_static} demonstrate a similar trend. Compared to LiDAR-NeRF, the hybrid representation and ray-drop refinement of LiDAR4D lead to a 12.0\% and 13.7\% drop in the depth and intensity RMSE metrics.

\begin{figure*}[th]
\centering
  \includegraphics[width=0.9\textwidth]{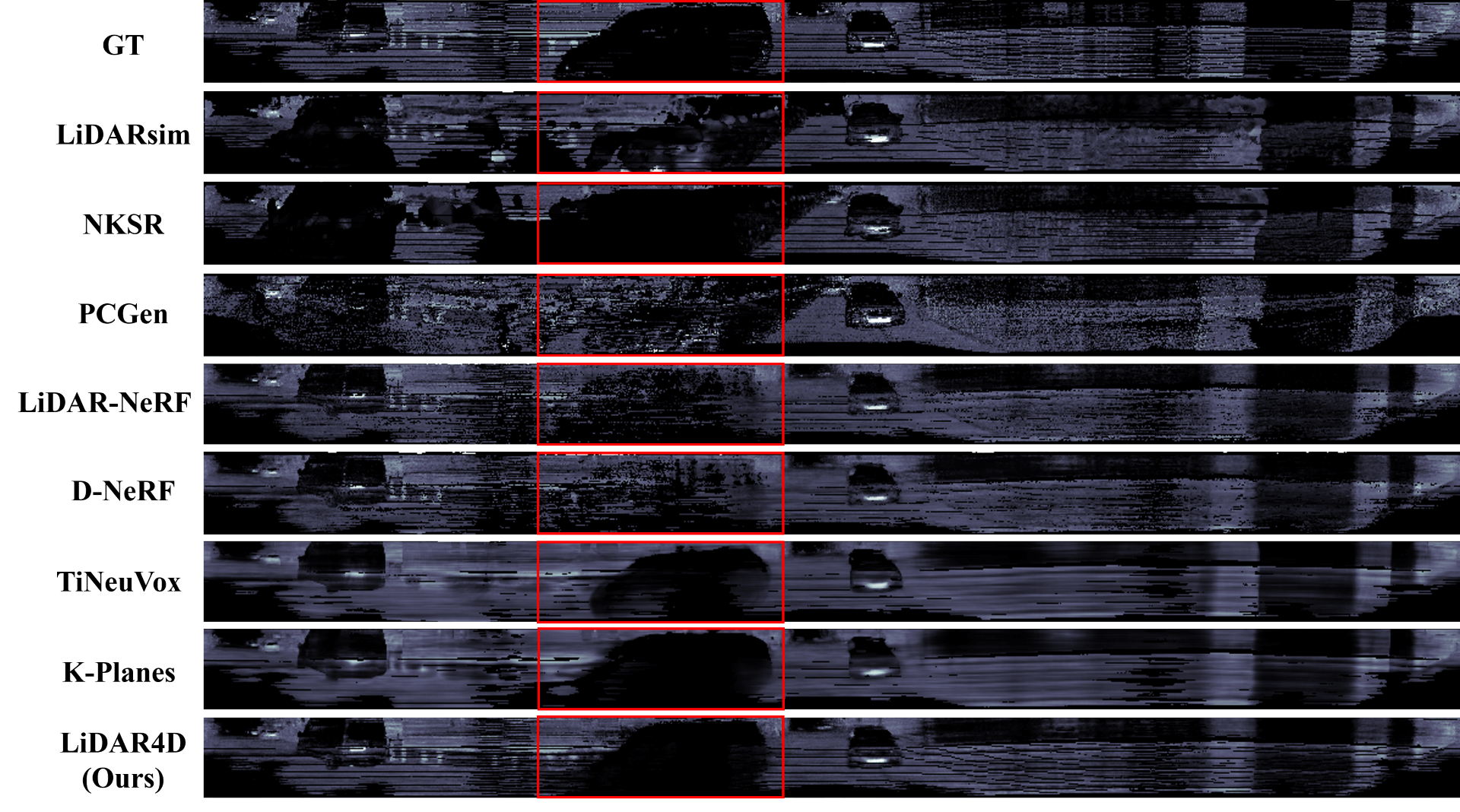}
  \caption{\textbf{Qualitative comparison for LiDAR intensity reconstruction and synthesis.} \textit{Dynamic} vehicles are marked with red boxes. }
  \label{fig:results_intensity}
\end{figure*}

\begin{figure*}[th]
\centering
  \includegraphics[width=0.9\textwidth]{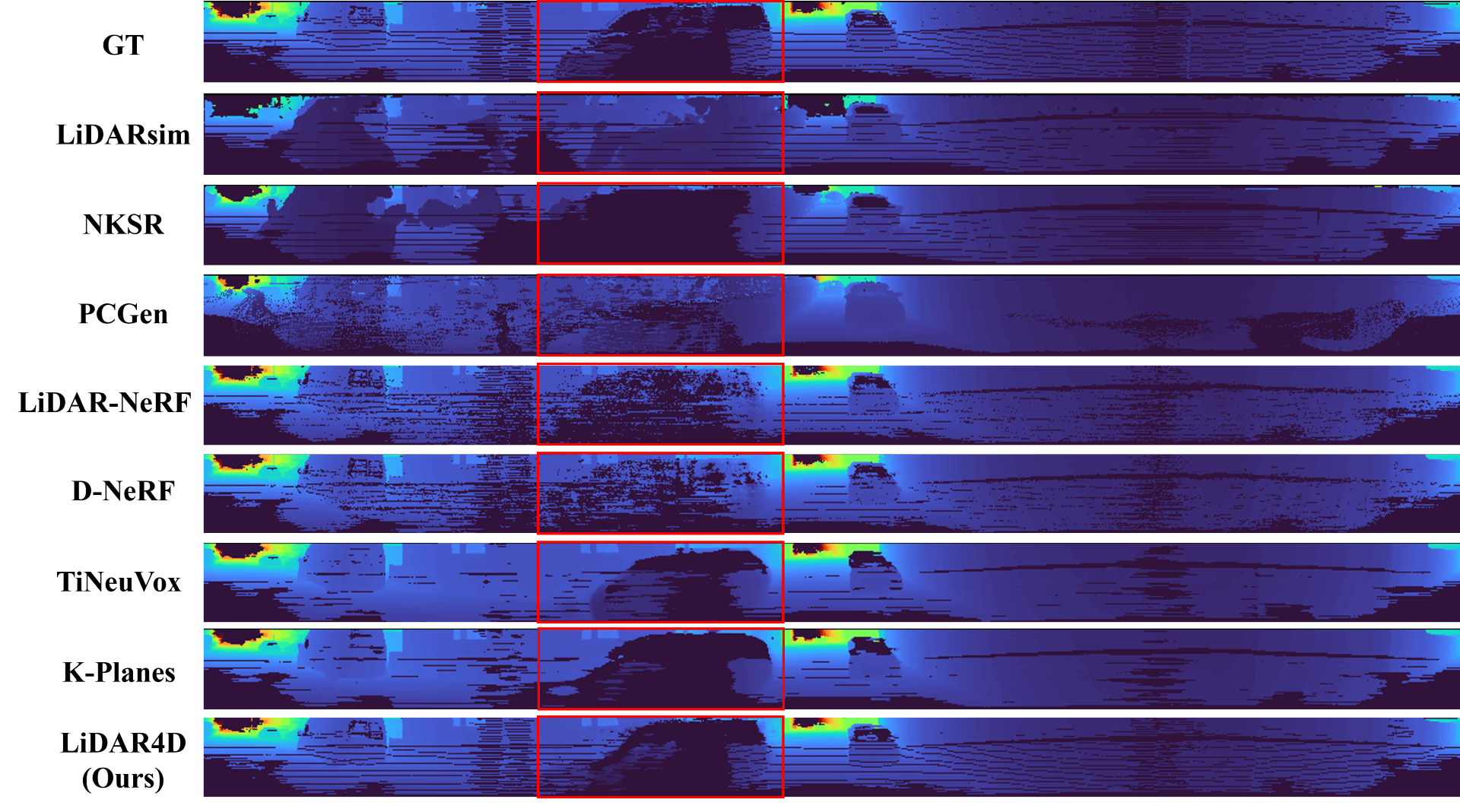}
  \caption{\textbf{Qualitative comparison for LiDAR depth reconstruction and synthesis.} \textit{Dynamic} vehicles are marked with red boxes.  }
  \label{fig:results_depth}
\end{figure*}

\noindent
\textbf{Dynamic Scenes.} Explicit reconstruction methods fail completely in dynamic scenes (\Cref{fig:IntensityRecon,fig:DepthRecon,fig:results_intensity,fig:results_depth}), which yields extremely poor validation results (\Cref{exp:kitti360_dynamic,exp:nuscenes}) due to the stacking of dynamic objects. In contrast, implicit reconstruction methods largely avoid the artifacts and noise of dynamic objects. However, existing methods like LiDAR-NeRF are designed for static scenes, resulting in the obscureness or absence of moving objects (\Cref{fig:pointcloud_vis,fig:pointcloud_nuscenes}). Although D-NeRF~\cite{pumarola2021dnerf} incorporates a deformation field, its impact is quite limited. The primary issue lies in the lack of constraints and the difficulty of establishing long-distance correspondence. Moreover, the state-of-the-art dynamic methods TiNeuVox~\cite{fang2022tineuvox} and K-planes~\cite{fridovich2023kplanes} are limited by their representation resolution, which makes it difficult to reconstruct details in large-scale scenes, such as vehicle and pedestrian geometry (\textit{Row 7\&8} in \Cref{fig:DepthRecon}), as well as high-frequency details in intensity (\textit{Row 7\&8} in \Cref{fig:results_intensity}). Our proposed LiDAR4D instead accomplishes geometric-aware and time-consistent dynamic reconstruction through 4D hybrid representation and flow-constrained temporal feature aggregation. As shown in~\Cref{exp:kitti360_dynamic,exp:nuscenes}, LiDAR4D ranks first across almost all metrics. A considerable visualization intuitively exhibits the superior generation quality of LiDAR4D, encompassing both long-distance moving vehicles and small bicyclists (the last row in~\Cref{fig:DepthRecon,fig:results_depth}).


\begin{table}[t]
\centering
\scalebox{0.75}{
\renewcommand{\arraystretch}{1.5}
\begin{tabular}{p{0.02\textwidth}p{0.12\textwidth}<{\centering}p{0.05\textwidth}<{\centering}p{0.05\textwidth}<{\centering}p{0.05\textwidth}<{\centering}p{0.05\textwidth}<{\centering}p{0.05\textwidth}<{\centering}p{0.05\textwidth}<{\centering}}
\hline \multicolumn{2}{c}{\multirow[c]{2}{*}{w/ GT Mask}} & \multicolumn{3}{c}{ Depth } & \multicolumn{3}{c}{ Intensity } \\
& & LPIPS$\downarrow$ & SSIM$\uparrow$ & PSNR$\uparrow$ & LPIPS$\downarrow$ & SSIM$\uparrow$ & PSNR$\uparrow$ \\
\hline
 
  \multirow[c]{2}{*}{$\mathcal{S.}$} & LiDAR-NeRF & 0.025 & 0.971 & 34.808 & 0.146 & 0.667 & 19.935 \\

   & Ours & \textbf{0.024} & \textbf{0.974} & \textbf{36.303} & \textbf{0.145} & \textbf{0.704} & \textbf{20.677} \\

\hline

  \multirow[c]{2}{*}{$\mathcal{D.}$}  & LiDAR-NeRF & 0.126 & 0.843 & 29.361 & 0.192 & 0.583 & 18.891 \\

   & Ours & \textbf{0.019} & \textbf{0.981} & \textbf{36.222} & \textbf{0.137} & \textbf{0.715} & \textbf{21.407}  \\

\hline
\end{tabular}}
\caption{\textbf{Experiments with GT ray-drop mask on KITTI-360 Dataset}. $\mathcal{S}$: Static sequences, $\mathcal{D}$: Dynamic sequences.} 
\vspace{.2cm}
\label{exp:gt_mask}
\end{table}

\noindent
\textbf{Difference on Ray-drop.} Existing methods differ in ray-drop modeling. PCGen~\cite{li2023pcgen} employs MLP to estimate ray-drop, while LiDARsim~\cite{manivasagam2020lidarsim} adopts U-Net, which takes depth and intensity values as input. In contrast, LiDAR4D predicts the ray drop probability of each point in space through neural fields and integrates them along the ray as the inputs of U-Net. Then, the U-Net is optimized in runtime to refine the prediction for individual scenarios. As can be seen from~\Cref{fig:results_static}, the MLP-based method may handle high-frequency details, but it also results in noisy prediction (\textit{Row 4\&5}). The U-Net-based method preserves global patterns better (\textit{Row 2\&3}) and consequently achieves superior results in LPIPS~\cite{zhang2018lpips} metrics in~\Cref{exp:kitti360_dynamic,exp:kitti360_static} in particular. However, this data-driven paradigm is dependent on the distribution of the training samples and is difficult to predict accurately in detail, \ie, the vehicle windows. LiDAR4D combines the advantages of both to achieve more realistic ray-drop modeling, as shown in~\Cref{fig:raydrop_vis}.

\subsection{Experiments without Ray-drop Effect}
In order to eliminate the effect of ray-drop on the evaluation metrics, we conduct supplementary experiments by only calculating the results on rays that have valid values. In other words, we apply the ground-truth ray-drop mask to all results for reconstruction quality evaluation. As shown in \Cref{exp:gt_mask}, our method outperforms LiDAR-NeRF~\cite{tao2023lidarnerf} in both static and dynamic scenarios, especially by a large margin in dynamic sequences.

\begin{figure}[t]
\centering
  \includegraphics[width=0.475\textwidth]{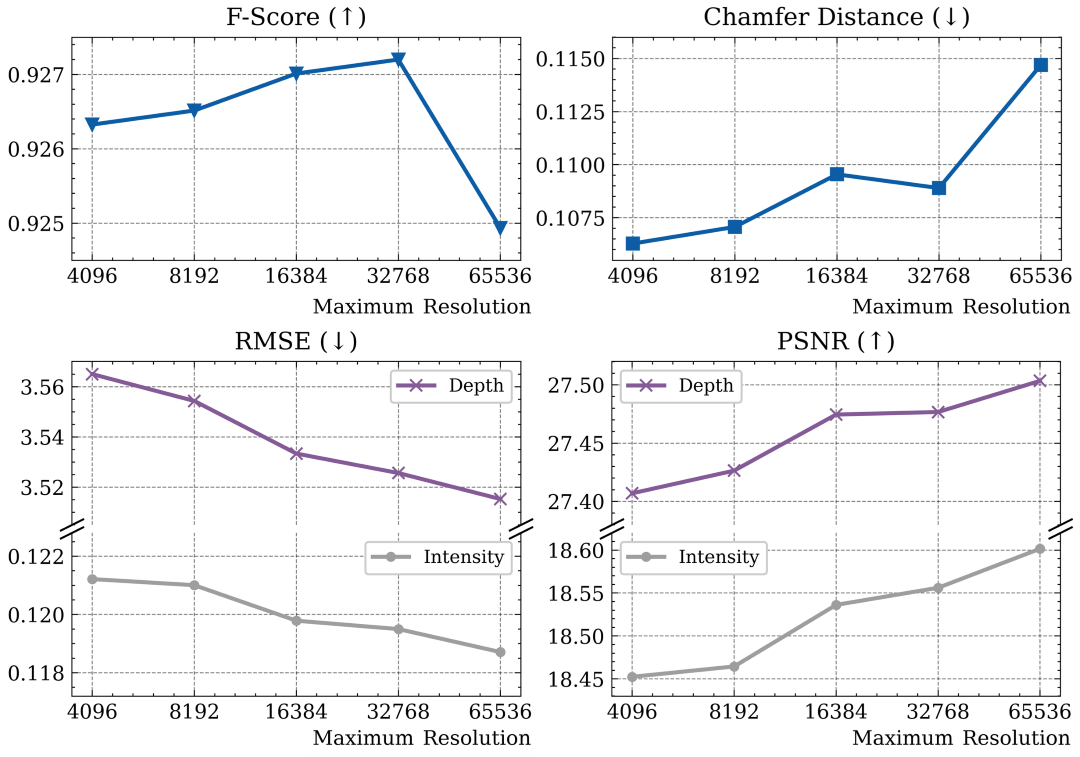}
  \vspace{-.4cm}
  \caption{\textbf{Influence of maximum representation resolution.} }
  \label{fig:resolution}
  \vspace{-.2cm}
\end{figure}  

\subsection{Experiments on Resolution}
Increasing the resolution of the representations is important for large-scale scenarios. In comparison to dense grids and planar features, hash grids can substantially raise the resolution and thus improve the accuracy of reconstruction, which has been verified in previous experiments. To determine the maximum resolution of hash grids, we further conducted additional experiments. As illustrated in~\Cref{fig:resolution}, increasing the resolution continuously alleviates the error of depth and intensity reconstruction. Considering the limited capacity, an extremely high resolution may lead to unfavorable effects, such as the degradation of point cloud metrics. Finally, we select the resolution of $2^{15}$, which can adequately meet the requirements of large-scale scene reconstruction.

\begin{figure*}[th]
\centering
  \includegraphics[width=0.9\textwidth]{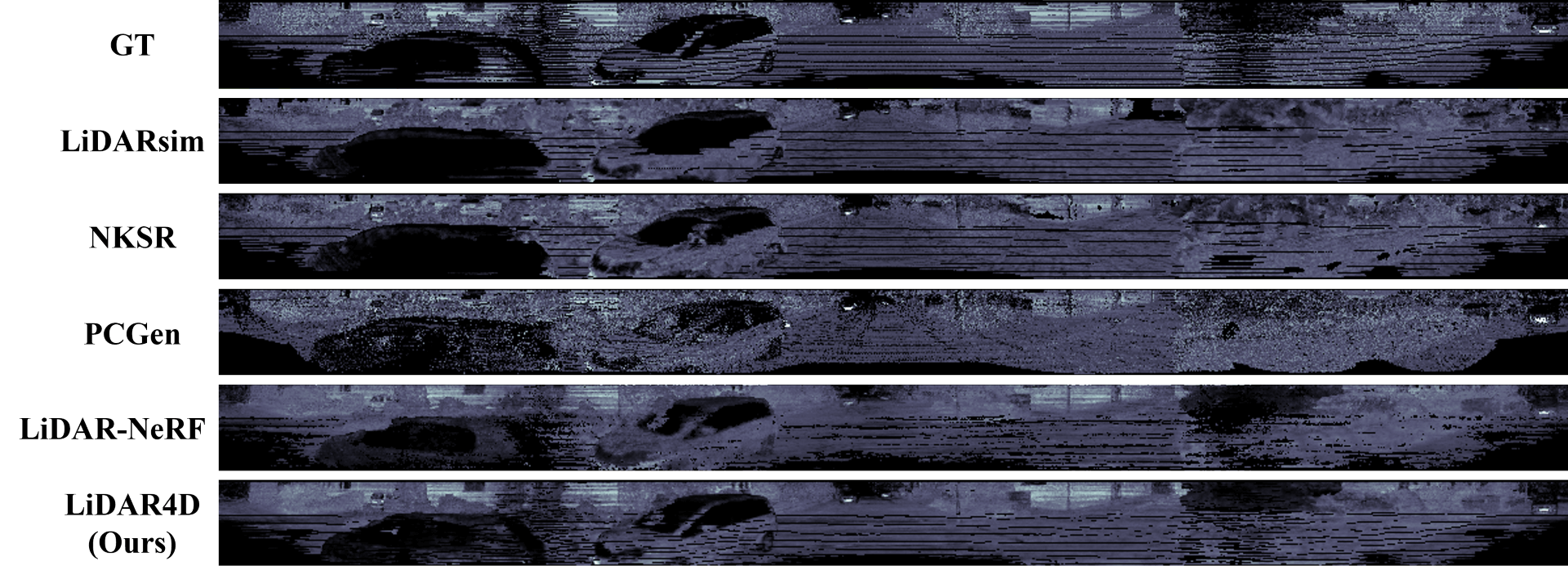}
\end{figure*} 

\begin{figure*}[th]
\centering
  \includegraphics[width=0.9\textwidth]{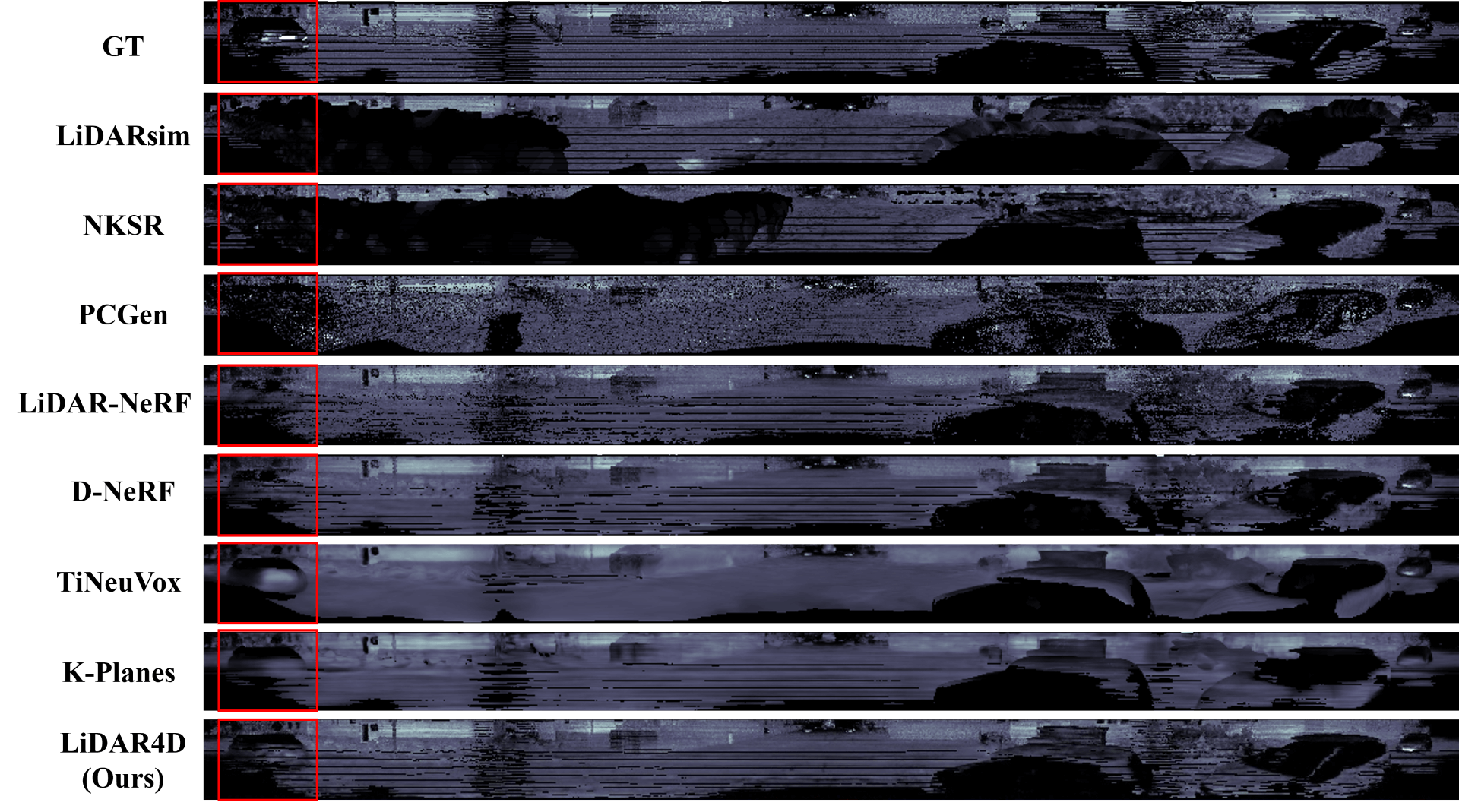}
\end{figure*}  

\begin{figure*}[th]
\centering
  \includegraphics[width=0.9\textwidth]{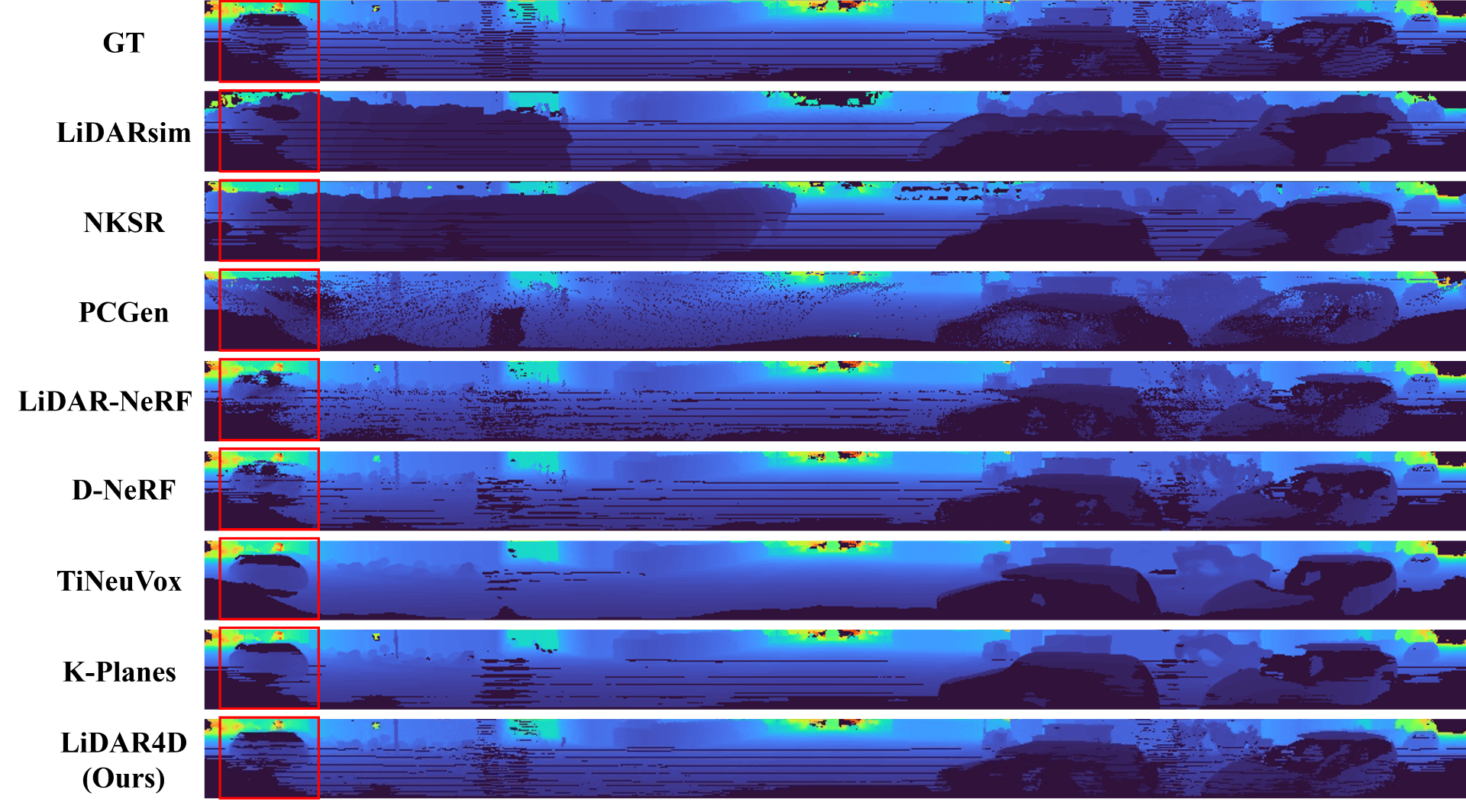}
\end{figure*} 

\begin{table}[]
\scalebox{0.75}{
\renewcommand{\arraystretch}{1.25}
\begin{tabular}{cccc}
\hline
\multicolumn{2}{c}{\textbf{KITTI-360}} & \multicolumn{2}{c}{\textbf{NuScenes}} \\ \hline
\multirow{4}{*}{\textbf{Static}} & Seq 1538-1601 & \multirow{10}{*}{\textbf{Dynamic}} & \multirow{2}{*}{Seq 450-500} \\
 & Seq 1728-1791 &  &  \\
 & Seq 1908-1971 &  & \multirow{2}{*}{\begin{tabular}[c]{@{}c@{}}Seq 1250-1300\\ (ego-vehicle stationary)\end{tabular}} \\
 & Seq 3353-3416 &  &  \\ \cline{1-2}
\multirow{6}{*}{\textbf{Dynamic}} & Seq 2350-2400 &  & \multirow{2}{*}{Seq 1600-1650} \\
 & Seq 4950-5000 &  &  \\
 & Seq 8120-8170 &  & \multirow{2}{*}{Seq 2200-2250} \\
 & Seq 10200-10250 &  &  \\
 & Seq 10750-10800 &  & \multirow{2}{*}{Seq 3180-3230} \\
 & Seq 11400-11450 &  & \\ \hline
\end{tabular}
}
\vspace{-.1cm}
\caption{\textbf{Scene sequences of KITTI-360 and NuScenes}.} 
\vspace{-.1cm}
\label{table:data_seq}
\end{table}

\section{Implementation Details}
\label{sec:implementation_details}

\subsection{Dataset Visualization}
\label{sec:dataset_visualization}
\begin{figure*}[ht]
\centering
  \includegraphics[width=0.98\textwidth]{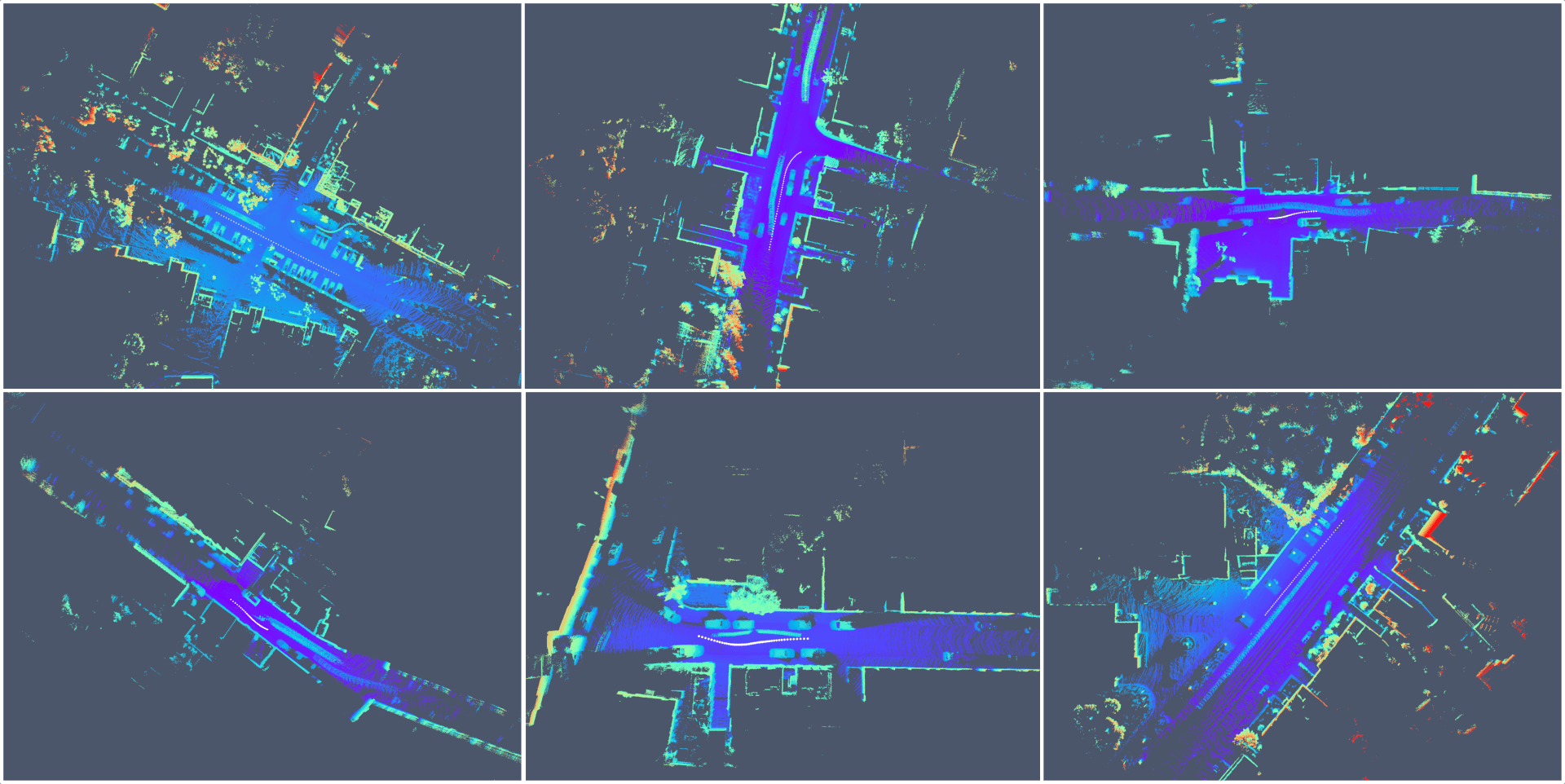}
  \vspace{-.2cm}
  \caption{\textbf{Visualization for the \textit{dynamic} sequences of KITTI-360 dataset.}}
  \label{fig:dataset_kitti}
\end{figure*}

\begin{figure*}[ht]
\centering
  \includegraphics[width=0.98\textwidth]{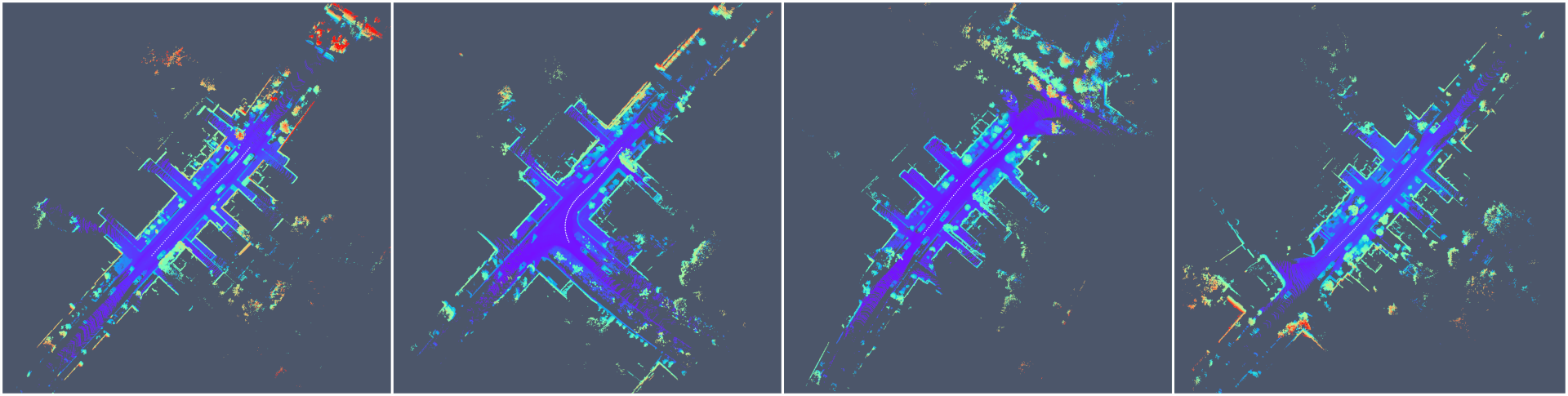}
  \vspace{-.2cm}
  \caption{\textbf{Visualization for the \textit{static} sequences of KITTI-360 dataset.}}
  \label{fig:dataset_kitti_static}
\end{figure*}

\begin{figure*}[ht]
\centering
  \includegraphics[width=0.98\textwidth]{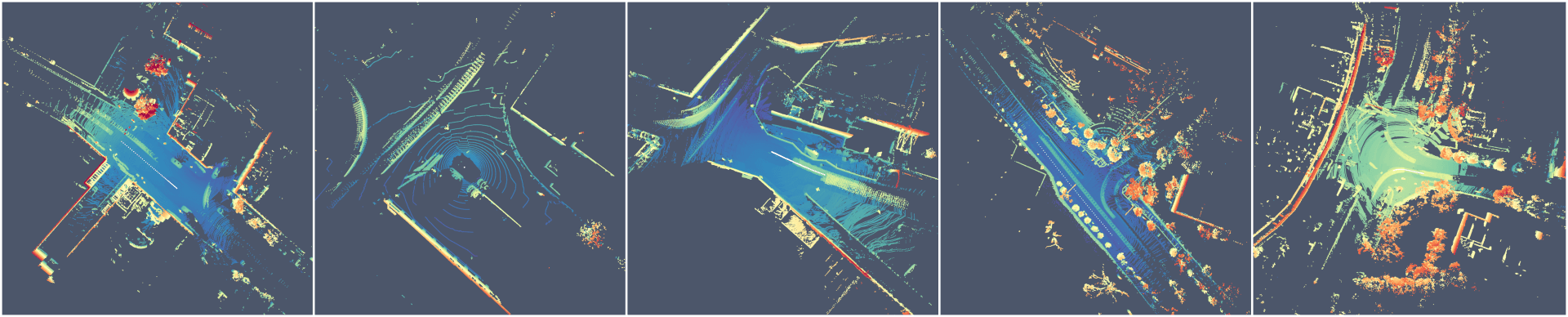}
  \vspace{-.2cm}
  \caption{\textbf{Visualization for the \textit{dynamic} sequences of NuScenes dataset.}}
  \label{fig:dataset_nuscenes}
\end{figure*}

As shown in~\Cref{fig:dataset_kitti}, we selected 6 representative dynamic scene sequences on KITTI-360. Each scene spans a distance of about 100–200 m and contains vehicles or pedestrians moving over long distances. Following the setup of LiDAR-NeRF~\cite{tao2023lidarnerf}, the same experiments were conducted on the original 4 static scene sequences (\Cref{fig:dataset_kitti_static}). The height and width of the range images are 66 and 1030. For the NuScenes dataset, we chose 5 dynamic sequences illustrated in~\Cref{fig:dataset_nuscenes}, including an ego-vehicle stationary scene (\textit{Column 2}) which can be viewed as a special case of novel temporal view synthesis. The size of the range images is set to 32 $\times$ 1080. The substantial variations between scenarios serve as a more accurate indicator of the reconstruction capabilities of current methods. The index number of scene sequences can be found in~\Cref{table:data_seq}.

\subsection{LiDAR4D}
\noindent
\textbf{Hybrid representation.} For the multi-planar features, the base resolution of the spatial plane is set to 64. The multi-scale structure has 4 levels, each doubling the spatial resolution and output 8-dimension feature, which finally yields a 32-dimension feature for both static and dynamic parts. The spatial resolution of hash grids ranges from 512 (the maximum resolution of multi-planar features) to $2^{15}$. There are a total of 8 levels of hash grids, each level outputs 4-dimension features, and then the same 32-dimension features are obtained. The grid is mapped to a hash table of $2^{19}$. All the temporal resolution is fixed to 25. Ultimately, the static and dynamic features of the planes and hash grids compose a 128-dimensional latent vector.

\vspace{0.1cm}

\noindent
\textbf{Dynamic modules.} Beyond time-conditioned multi-planar and hash grid features for dynamic reconstruction, we additionally introduce flow MLP to aggregate dynamic features for temporal consistency. This coordinate-based MLP is an 8-layer neural field with 128 units per layer. Eventually, the dynamic features of adjacent spatio-temportal points are aggregated by weighted averaging. We incorporate the chamfer distance loss based on point clouds to effectively constrain the optimization of the flow MLP. It encourages the two adjacent frames of the point cloud transformed by the predicted scene flow to be as consistent as possible. According to the training process, we randomly select one moment in each epoch for optimization. In addition, we preprocess the point cloud by removing ground points using RANSAC~\cite{fischler1981ransac} and further limiting the maximum distance within 50 meters to mitigate the adverse effect of noise.

\begin{figure*}[th]
\centering
  \includegraphics[width=0.97\textwidth]{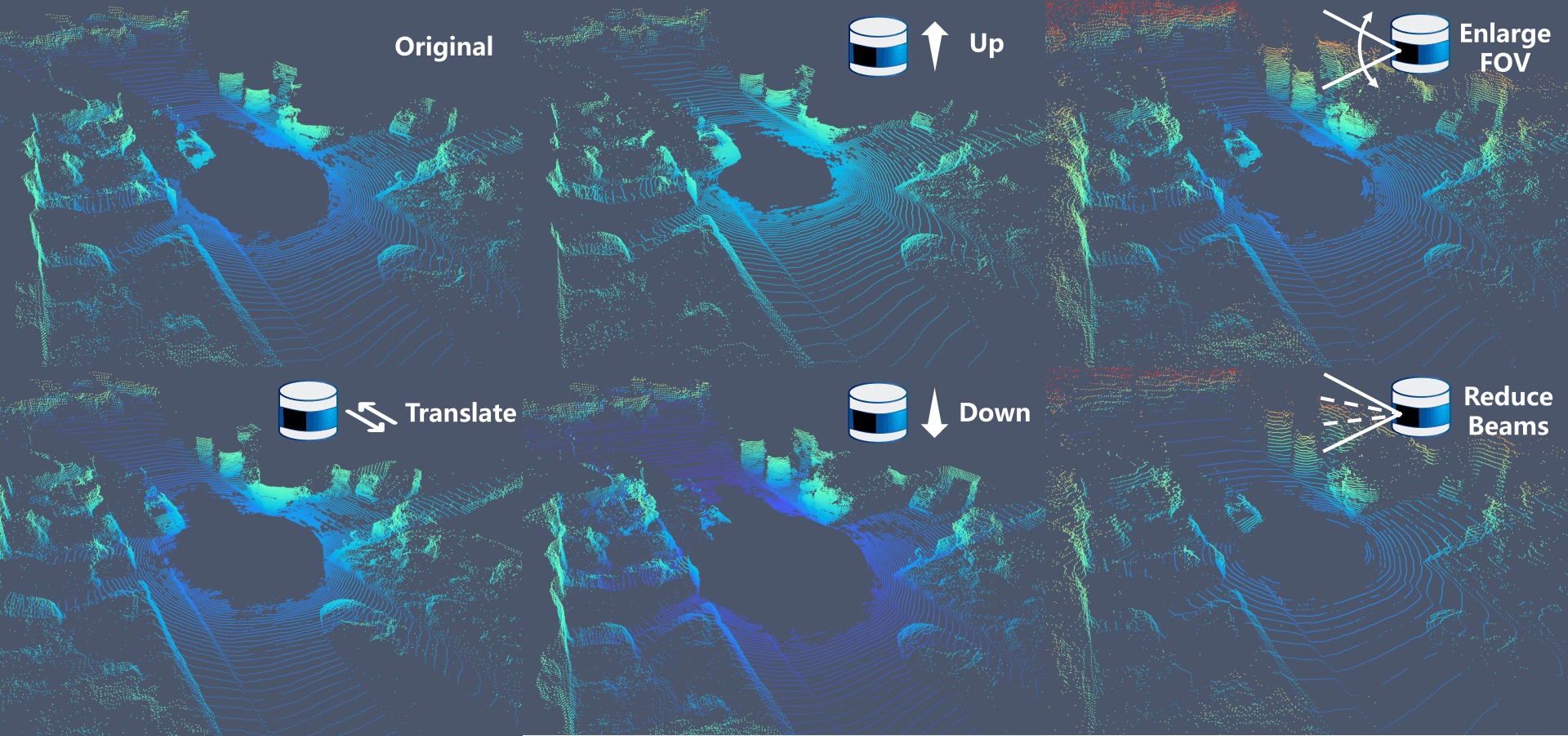}
  \vspace{-.2cm}
  \caption{\textbf{Novel view point cloud synthesis with different LiDAR configurations.}}
  \label{fig:application}
\vspace{-.1cm}
\end{figure*} 

\begin{figure*}[th]
\centering
  \includegraphics[width=0.97\textwidth]{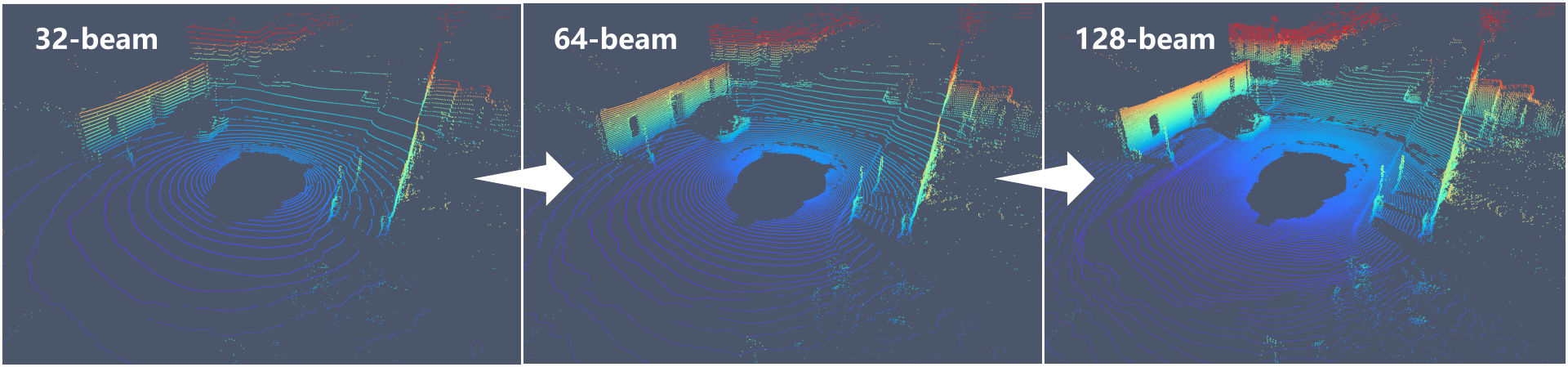}
  \vspace{-.2cm}
  \caption{\textbf{Increase LiDAR beams to densify the point cloud on NuScenes dataset.}}
  \label{fig:application_dense}
\vspace{-.1cm}
\end{figure*} 

\vspace{0.3cm}

\noindent
\textbf{Neural LiDAR fields.} The aggregated time-conditioned and flow-constrained dynamic features are finally fed into a 2-layer 64-dimensional MLP, which outputs the 15-dimensional geometric feature and density value. The geometric feature with the 12-band frequency-encoded viewpoint is utilized to predict intensity values and ray-drop probabilities by two independent 3-layer 64-dimensional MLPs, respectively. The expectation of the density integrated along the ray serves as the depth value. Then, these initial predictions are combined as inputs to U-Net for global ray-drop optimization. The final predictions are multiplied by the ray-drop mask for synthesis.

\noindent
\textbf{Optimization details.} The initial learning rate is set to 0.01 for the multi-planar and hash grids, and 0.001 for other MLP networks. The learning rate decreases exponentially during iterations, with a final decay coefficient of 0.1. The depth-loss weight $\lambda_{\alpha}$ is 1, the intensity weight $\lambda_{\beta}$ is 0.1, the ray-drop weight $\lambda_{\gamma}$ is 0.01, and the flow weight $\lambda_{\eta}$ is 0.01. During refinement, the weight $\lambda_\mathrm{r}$ is set to 1 with other loss weights set to 0. The U-Net weights are randomly initialized and optimized with a learning rate of 0.001 by the Adam~\cite{kingma2014adam} optimizer. Other unmentioned optimization details are basically in line with LiDAR-NeRF~\cite{tao2023lidarnerf}.

\noindent
\textbf{Efficiency.} According to experiments conducted on a single NVIDIA GeForce RTX 4090 GPU, LiDAR4D takes about 2 hours to complete the optimization of each scenario.

\section{Applications}
\label{sec:applications}
At last, we showcase the application of LiDAR4D for novel-view LiDAR synthesis with different sensor configurations. As illustrated in~\Cref{fig:application}, we can freely manipulate the sensor's pose, \eg, moving up and down or horizontal translation. It can be observed that the LiDAR point clouds under different sensor poses vary significantly, and the accurate recovery of the scene and objects further demonstrates the high-fidelity synthesis of LiDAR4D. In addition, we can adjust LiDAR configurations, such as increasing the vertical field of view, to obtain a wider range of sensing results on the top right of~\Cref{fig:application}. Alternately, the modification of LiDAR beams realizes the conversion between sparse and dense point clouds. As shown on the bottom right of~\Cref{fig:application}, we transfer the LiDAR configuration of KITTI-360 to that of NuScenes, realizing the crossing of the domain gap. Also as shown in~\Cref{fig:application_dense}, we can also densify the sparse Nuscenes data, which will also be beneficial for downstream tasks. All of this reveals the adaptability and enormous potential of LiDAR4D.

\end{document}